\documentclass[11pt,a4paper]{article}
\usepackage{times,latexsym}
\usepackage{url}
\usepackage[T1]{fontenc}

    \usepackage[acceptedWithA]{tacl2021v1}
\usepackage{xspace,mfirstuc,tabulary}

\newif\iftaclinstructions
\taclinstructionsfalse 
\iftaclinstructions
\renewcommand{\confidential}{}
\renewcommand{\anonsubtext}{(No author info supplied here, for consistency with
TACL-submission anonymization requirements)}
\newcommand{\instr}
\fi

\iftaclpubformat 

\else

\fi

\usepackage{times}
\usepackage{latexsym}
\usepackage{graphicx} 
\usepackage{amsfonts, amsmath}
\usepackage{xcolor,colortbl}
\usepackage{multicol}
\usepackage{multirow}
\usepackage{tablefootnote}
\usepackage{longtable}
\usepackage{hyperref}

\usepackage{pgfplots}
\usepackage{blindtext, subfig}

\usepackage{algorithm}
\usepackage{algpseudocode}
\usepackage{multirow}
\usepackage{enumitem}
\usepackage{booktabs}
\usepackage{ulem}
\pgfplotsset{width=7cm,compat=1.3}

\definecolor{darkgray2}{rgb}{0.36, 0.36, 0.36}
\definecolor{LightCyan}{rgb}{0.8,0.9,0.8}
\definecolor{Gray}{gray}{0.93}

\newcommand\Tstrut{\rule{0pt}{2.6ex}}       
\newcommand\Bstrut{\rule[-0.9ex]{0pt}{0pt}} 
\newcommand{\TBstrut}{\Tstrut\Bstrut} 
\newcommand{\X}{FaRM}
\newcommand{\FARM}{\textbf{Fa}irness-aware \textbf{R}ate \textbf{M}aximization}
\newcommand{\dial}{\textsc{Dial}}
\newcommand{\pan}{\textsc{Pan16}}
\newcommand*{\rowstyle}[1]{
  \gdef\@rowstyle{#1}%
  \@rowstyle\ignorespaces%
}
\makeatletter
\newcommand{\thefontsize}{The current font size is: \f@size pt}

\usepackage[T1]{fontenc}

\usepackage[utf8]{inputenc}

\usepackage{microtype}

\makeatletter
\newcommand{\ssymbol}[1]{^{\@fnsymbol{#1}}}

\usepackage[linecolor=orange]{todonotes}

\newcommand{\SBB}[1]{\textcolor{black}{#1}}

\definecolor{navyblue}{RGB}{102, 178, 255}
\definecolor{green}{RGB}{0, 153, 0}

\title{Learning Fair Representations via Rate-Distortion Maximization}

\author{
Somnath Basu Roy Chowdhury\qquad Snigdha Chaturvedi \\
  \texttt{\{somnath, snigdha\}@cs.unc.edu} \\
  UNC Chapel Hill \\
}

\date{}

\begin{document}
\maketitle
\begin{abstract}

Text representations learned by machine learning models  often encode undesirable demographic information of the user. 
Predictive models {based on these representations can rely on} such information, resulting in biased decisions. We present a novel debiasing technique, {\FARM} ({\X}), that removes protected information by making representations of instances belonging to the same protected attribute class uncorrelated, using the rate-distortion function. {\X} is able to debias representations with or without a target task at hand. {\X} can also be adapted to remove information about multiple protected attributes simultaneously.  Empirical evaluations show that {\X} achieves state-of-the-art performance on several datasets, and learned representations leak significantly less protected attribute information against an attack by a non-linear probing network.
\end{abstract}

\section{Introduction}
Democratization of machine learning has led to deployment of 
predictive models for critical  applications like credit approval \cite{ghailan2016improving} and college application reviewing \cite{basu2019predictive}. 
Therefore, 
it is important to ensure that decisions made by {these} models are \textit{fair} towards different demographic groups \cite{mehrabi2019survey}. 
{Fairness can be achieved} 
by ensuring that the demographic information does not get encoded in {the representations used by these 
models} \cite{blodgett2016demographic, elazar2018adversarial, elazar2021amnesic}.

{However, } controlling demographic information encoded in a model's representations is a challenging task for textual data. {This is because} natural language text is highly indicative of an author's demographic attributes even when it is not explicitly mentioned \cite{koppel2002automatically, burger2011discriminating, nguyen2013old, verhoeven2014clips, weren2014examining, rangel2016overview, verhoeven2016twisty, blodgett2016demographic}.

\begin{figure}[t!]
    \centering
    \includegraphics[width=0.48\textwidth, keepaspectratio]{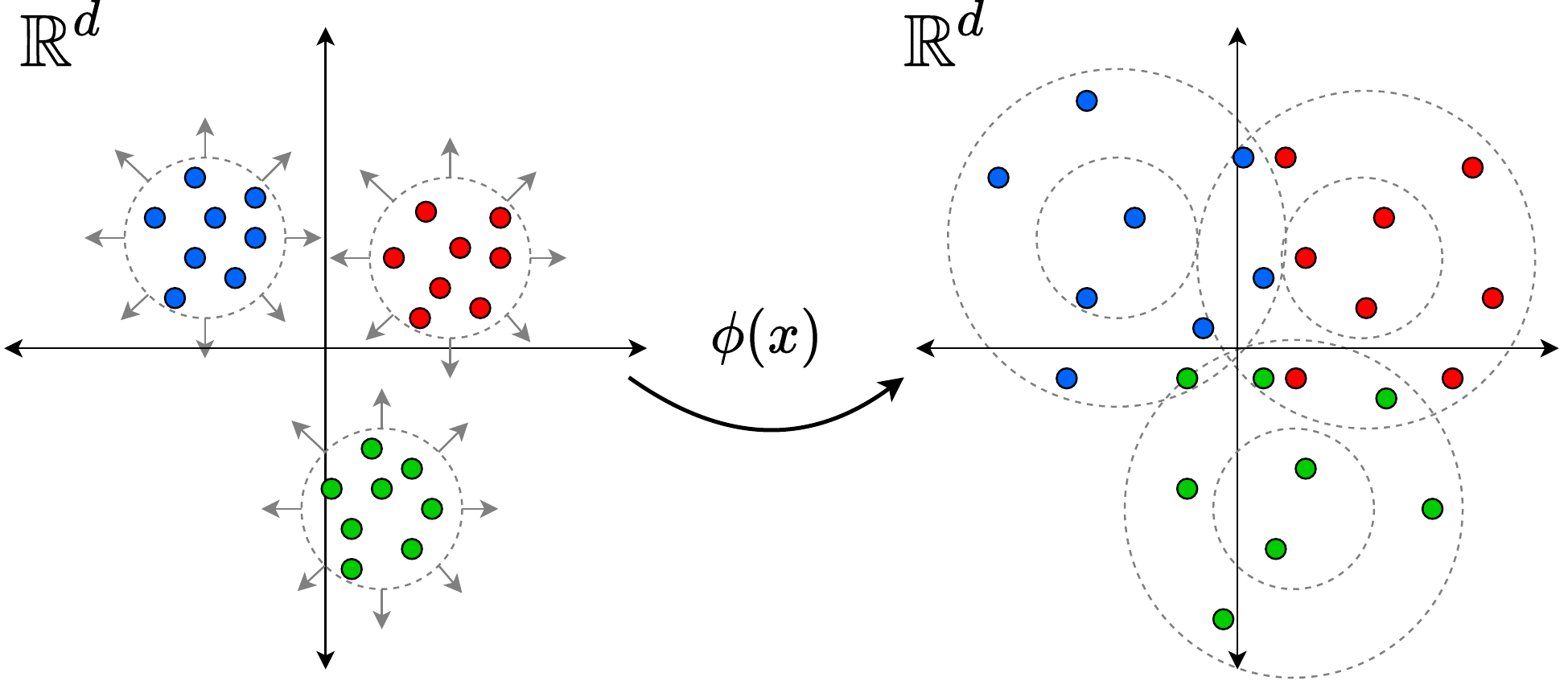}
    \caption{\small Illustration of {unconstrained} debiasing using {\X}. 
    Representations are color-coded (in blue, red and green) according to their protected attribute class. 
    Before debiasing (left), representations within each class are similar to each other (intra-class information content is low). Debiasing 
    enforces the within class representations to be uncorrelated by increasing their information content.}
    \label{fig:intuition}
\end{figure}

In this work, we 
\textit{debias} information about a protected 
attribute (e.g.\ gender, race) 
from textual data representations. Previous debiasing methods \cite{bolukbasi2016man, ravfogel2020null} project representations in a subspace that does not reveal protected attribute information. These methods are only able to guard protected attributes against an attack by a linear function~\cite{ravfogel2020null}.
Other methods \cite{xie2017controllable, basu-roy-chowdhury-etal-2021-adversarial} adversarially remove protected information while retaining information about a target attribute. {However, } 
they are 
{difficult} to train \cite{elazar2018adversarial} and require 
a target task at hand.



We present a novel debiasing technique, {\FARM} (\X), that removes demographic information by controlling the \textit{rate-distortion} function of the learned representations. Intuitively, in order to remove information about a protected attribute {from a set of representations}, we want the 
representations from the same {protected} attribute class to be {uncorrelated} to each other. We achieve this by maximizing the {number of bits} (rate-distortion) required to encode  representations 
with the same protected attribute.
Figure~\ref{fig:intuition} illustrates the process. The representations are shown as points in a two-dimensional feature space, color-coded according to their protected attribute class. {\X} learns a function $\phi(x)$ such that representations of the same protected class become uncorrelated and similar to other representations, thereby making it difficult to extract the {information about the protected attribute from the learned representations.} 


We perform rate-distortion maximization based debiasing in the following setups: (a) \textit{unconstrained debiasing} -- we remove information about a protected attribute $\mathbf g$ while retaining remaining information as much as possible \SBB{(e.g. debiasing gender information from word embeddings)}, 
and (b) \textit{constrained debiasing} -- we retain information about a target attribute $\mathbf y$ while removing information pertaining to $\mathbf g$ \SBB{(e.g. removing racial information from representations during text classification). In the unconstrained setup, debiased representations can be used for different downstream tasks, whereas for constrained debiasing the user is interested only in the target task}. For unconstrained debiasing, we evaluate {\X} for removing gender information from word embeddings and demographic information from text representations that can then be used for a downstream NLP task (we show their utility for biography and sentiment classification in our experiments). 
Our empirical evaluations show that representations learned using {\X} in an unconstrained setup leak significantly less protected attribute information compared to prior approaches against an attack by a non-linear probing network. 

For constrained debiasing, 
{\X} achieves state-of-the-art debiasing performance on 3 datasets, and representations are able to guard protected attribute information significantly better than previous approaches. We also perform experiments to show that {\X} is able to remove multiple {protected} attributes simultaneously while guarding against intersectional group  biases \cite{subramanian2021evaluating}. To summarize, our main contributions are: 

\begin{itemize}[noitemsep, topsep=0.3pt, leftmargin=*]
    \setlength\itemsep{0.3em}
    \item We present 
    {\FARM} (\X) for  debiasing of textual data representations in {unconstrained} and {constrained} setups, by controlling their {rate-distortion} functions.
    \item 
    We empirically show {\X} leaks significantly less protected 
    information against a {non-linear probing attack}, outperforming prior approaches.
    \item We present two variations of {\X} for debiasing multiple {protected} attributes simultaneously, which is also effective against an attack for intersectional group biases.
\end{itemize}

\section{Related Work}

Removing sensitive 
attributes {from data representations} for fair classification was initially introduced as an optimization task 
\cite{zemel2013learning}. {Subsequent works have used adversarial frameworks \cite{goodfellow2014generative} for this task} 
\cite{zhang2018mitigating, li2018towards, xie2017controllable, elazar2018adversarial, basu-roy-chowdhury-etal-2021-adversarial}. {However,} adversarial networks are difficult to train \cite{elazar2018adversarial} and cannot function  without a target task at hand. 

Unconstrained debiasing frameworks focus on removing a protected attribute from representations, without relying on a target task.  \citet{bolukbasi2016man} demonstrated that GloVe embeddings encode gender information, and proposed an unconstrained debiasing framework {for identifying gender direction} and neutralizing vectors along that direction. Building on this approach, \citet{ravfogel2020null} proposed INLP, a robust framework to debias representations 
by iteratively identifying protected attribute subspaces and projecting representations onto the corresponding nullspaces. \SBB{However, these approaches fail to guard protected information against an attack by a non-linear probing network. \citet{dev2020oscar} showcased that nullspace projection approaches can be extended for debiasing in a constrained setup as well.} 
 
 In contrast to prior works, we present a novel debiasing framework based on the principle of rate-distortion maximization. Coding rate maximization was introduced as an objective function by \citet{ma2007segmentation} for image segmentation. 
 {It has also been used in explaining feature selection by deep networks \cite{macdonald2019rate}.}
 Recently, \citet{yu2020learning} proposed {maximal coding rate} (MCR\textsuperscript{2}) based on rate-distortion theory, a representation-level objective function that can serve as an alternative to empirical risk minimization methods. Our work is similar to MCR\textsuperscript{2} as we learn representations using a rate-distortion framework, but instead of tuning representations for classification we remove protected attribute information from them. 

\section{Preliminaries}

\SBB{Our framework performs debiasing by making representations of the same protected attribute class uncorrelated. To achieve this, we leverage a principled objective function called rate-distortion, to measure the compactness of a set of representations.}
In this section, we introduce the fundamentals of rate-distortion {theory}.\footnote{\SBB{We borrow some notations from \citet{yu2020learning} to explain concepts of rate-distortion theory.}}

\noindent\textbf{Rate Distortion}. In lossy data compression \cite{cover1999elements}, the compactness of a random distribution is measured by the minimal number of binary bits required to encode it. 
A lossy coding scheme encodes a finite set of vectors  $Z = \{z_1, \ldots, z_n\} \in \mathbb{R}^{n \times d}$ from a distribution $P(Z)$, such that the decoded vectors $\{\hat{z}_i\}^n_{i=1}$ can be recovered up to a precision $\epsilon^2$.  
The \textit{rate-distortion} function $R(Z, \epsilon)$ 
measures the minimal number of bits per vector required to encode the sequence $Z$.

In case the vectors $\{z_i\}^n_{i=1}$ are i.i.d. samples from a zero-mean multi-dimensional Gaussian distribution $\mathcal{N}(0, \Sigma)$, the optimal rate-distortion function is given as:
\begin{equation}
    R(Z, \epsilon) = \frac{1}{2} \log_2 \det \left( I + \frac{d}{n\epsilon^2}ZZ^T\right)
    \label{eqn:rate}
\end{equation}

\noindent where $\frac{1}{n}ZZ^T=\hat{\Sigma}$ is the estimate of covariance matrix $\Sigma$ for the Gaussian distribution. \SBB{As the eigenvalues of the matrices $ZZ^T$ and $Z^TZ$ are equal, the rate-distortion function $R(Z, \epsilon)$ is the same for both of them \cite{ma2007segmentation}. In most setups $d \ll n$, therefore we use $Z^TZ$ for efficiently computing $R(Z, \epsilon)$.}

In rate-distortion theory, we need $nR(Z, \epsilon)$ bits to encode $n$ vectors of $Z$. The optimal codebook also depends on data dimension ($d$) and requires $dR(Z, \epsilon)$ bits to encode. Therefore, 
a total of $(n + d)R(Z, \epsilon)$ is bits required to encode the sequence $Z$.
{\citet{ma2007segmentation} showed that this provides a tight bound 
even in cases where the underlying distribution $P(Z)$ is degenerate.} \SBB{This enables the use of this loss function for real-world data, where the underlying distribution may not be well defined.}

{In general, a set of compact vectors 
(low information content) would require a small number of bits to encode, which would correspond to a small value of  $R(Z, \epsilon)$ and vice versa. }

\noindent\textbf{Rate Distortion for a mixed distribution}. In general, the set of vectors $Z$ can be from a mixture distribution (e.g. feature representations for multi-class data). The rate-distortion function can be computed {by splitting} the data into  multiple subsets: $Z = Z^1 \cup Z^2 \ldots \cup Z^k$, based on their distribution. For each subset, we can compute the $R(Z^i, \epsilon)$ (Equation~\ref{eqn:rate}). To facilitate the computation, we define a membership matrix $\Pi = \{\Pi_j\}^k_{j=1}$ as a set of $k$ matrices to encode membership information in each subset $Z^j$. The membership matrix $\Pi_j$ for each subset is a diagonal matrix defined as: 
\begin{equation}
    \Pi_j = \mathrm{diag}(\pi_{1j}, \pi_{2j}, \ldots, \pi_{nj}) \in \mathbb{R}^{n \times n}
    \label{eqn:partition}
\end{equation}

\noindent where $\pi_{ij} \in [0, 1]$ denotes the probability of a vector $z_i$ belonging to the $j$-{th} subset and $n$ is the number of vectors in the sequence $Z$. The matrices satisfy the constraints: $\sum_j\pi_{ij} = 1,\; \sum_j \Pi_j = I_{n \times n},\; \Pi_j \succeq 0$.
\SBB{The expected number of vectors in the $j$-{th} subset $Z^j$ is $\mathrm{tr}(\Pi_j)$ and the corresponding covariance matrix: $\frac{1}{\mathrm{tr}(\Pi_j)} Z\Pi_j Z^T$. The overall rate-distortion function {is given as:}}

\begin{equation*}
    \begin{aligned}
    R^c(&Z, \epsilon | \Pi)  =\\ & \sum\limits_{j=1}^k \frac{\mathrm{tr}(\Pi_j) }{2n} \log_2 \det \left(I + \frac{d}{\mathrm{tr}(\Pi_j)\epsilon^2}Z\Pi_jZ^T\right)
    \end{aligned}
\end{equation*}

For multi-class data, where a vector $z_i$ can only be a member of a single class, {we restrict} $\pi_{ij} = \{0, 1\}$, and therefore the covariance matrix for $j$-th subset is ${Z^j}{Z^j}^T$. {In general, if the representations within each subset $Z^j$ are similar to each other, they will have low intra-class variance, and it would correspond to a small $R^c(Z, \epsilon|\Pi)$ and vice versa.}


\section{Fairness-Aware Rate Maximization}
In this section, we describe { \X} to debias representations in {unconstrained} and {constrained} setups.

\subsection{Unconstrained Debiasing using {\X}} 
In this setup, we aim to remove information about a protected attribute $\mathbf g$ from data representations $X$ while retaining the remaining information. To achieve this, the debiased representations $Z$ should have the following properties:\\
(a) \textit{Intra-class Incoherence}:  Representations belonging to the same protected attribute class should be highly {uncorrelated}. This would make it difficult for a classifier to extract  any information about $\mathbf g$ {from the representations}.\\
(b) \textit{Maximal Informativeness}: Representations should be maximally informative about the remaining information. 

Assuming there are $k$ protected attribute classes, we can write $Z = Z^1 \cup \ldots \cup Z^k$.
To achieve (a), we need to ensure that the representations in a subset $Z^j$ belonging to the same protected class are dissimilar 
 {and have large intra-class variance.
An increased intra-class variance would correspond to an increase in the number of bits to encode samples within each class and the rate-distortion function $R^c(Z, \epsilon|\Pi^{\mathbf{g}})$ would be \textit{large}. 
For (b), we want the representations $Z$ to retain maximal possible information from the input $X$. 
{Increasing information content in $Z$, would require a larger number of bits to encode it. This means that the rate-distortion $R(Z, \epsilon)$ should also be \textit{large}.} }

{{\X} achieves (a) and (b) simultaneously by \textit{maximizing} the following objective function: } 
\begin{equation}
    {J}_u(Z, \Pi^{\mathbf{g}}) = R^c(Z, \epsilon | \Pi^{\mathbf{g}}) + R(Z, \epsilon)
\end{equation}
where the membership matrix $\Pi^{\mathbf{g}}$, is constructed using the protected attribute $\mathbf g$ (see Equation~\ref{eqn:partition}). 



\begin{algorithm}[t!]
\SBB{
\caption{Unconstrained Debiasing Routine}
\label{alg:unconstrained}
\begin{algorithmic}[1]
    \State \textbf{Input}: $(X, G)$ {input data set with protected attribute labels. Number of training epochs $N$.}
    \For{$i = 1, \ldots, N $}
        \State $Z = \text{LayerNorm}(\phi(X))$ 
        \State $\Pi^{\mathbf g} = \text{ConstructMatrix}(G)$ 
        \Comment{retrieve membership matrix using $G$}
        \State Update $\phi$ using gradients $\mathop{\nabla}\limits_{\phi} J_u(Z, \Pi^{\mathbf g})$
    \EndFor
\State $Z_{\text{debiased}} = \phi(X)$ \Comment{debiased representations}
\State \Return $\phi$ \Comment{debiasing network}
\end{algorithmic}}
\end{algorithm}

\SBB{The unconstrained debiasing routine is described in Algorithm~\ref{alg:unconstrained}.}
We use a deep neural network $\phi$ as our feature map to obtain debiased representations $z = \phi(x)$. The objective function $J_u$ is sensitive to the scale of the representations. Therefore, we normalize the Frobenius norm of the representations to ensure individual input samples have an equal impact on the loss. We use layer normalization~\cite{ba2016layer} to ensure that all representations have the same magnitude and lie on a sphere 
$z_i \in \mathbb{S}^{d-1}(r)$ of radius $r$. \SBB{The feature encoder $\phi$ is updated using gradients from the objective function $J_u$. The debiased representations are retrieved by feeding input data $X$ through the trained network $\phi$.} An illustration of the debiasing process in the unconstrained setup is shown in Figure~\ref{fig:intuition}.


\begin{figure}[t!]
    \centering
    \includegraphics[width=0.3\textwidth, keepaspectratio]{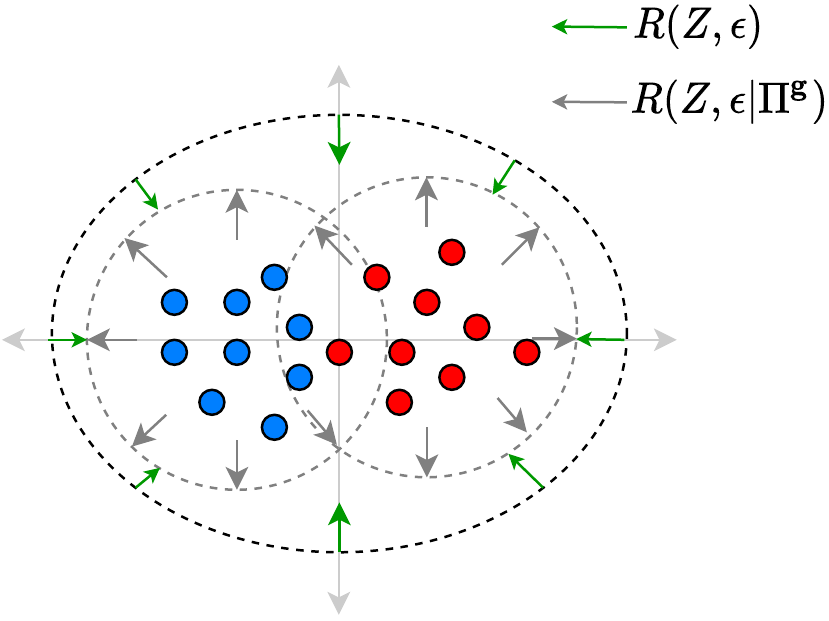}
    \caption{\small Visualization for regularization loss in $J_c$ for constrained debiasing. The \textcolor{red}{\textbf{red}} and \textcolor{blue}{\textbf{blue}} circles represent 2D representations from two different protected class. The \textcolor{gray}{\textbf{gray}} arrows are induced by $R^c(Z, \epsilon|\Pi^{\mathbf g})$ term and the \textcolor{green}{\textbf{green}} ones are induced by $R(Z, \epsilon)$ term.}
    \label{fig:constrained-viz}
    \vspace{-13pt}
\end{figure}

\subsection{Constrained Debiasing {using \X}} 
In this setup, we aim to {remove information about a protected attribute $\mathbf g$ from data representations $X$ while retaining information about a specific target attribute $\mathbf y$.} 
The learned representations should have the following properties:

\noindent(a) \textit{Target-Class Informativeness}: Representations should be maximally informative about the target task attribute $\mathbf y$. \\
\noindent(b) \textit{Inter-class Coherence}: Representations from different protected attribute classes should be \textit{similar} to each other. {This would make it difficult} to extract information about $\mathbf g$ from $Z$. 

\begin{figure}[t!]
    \centering
    \includegraphics[width=0.36\textwidth, keepaspectratio]{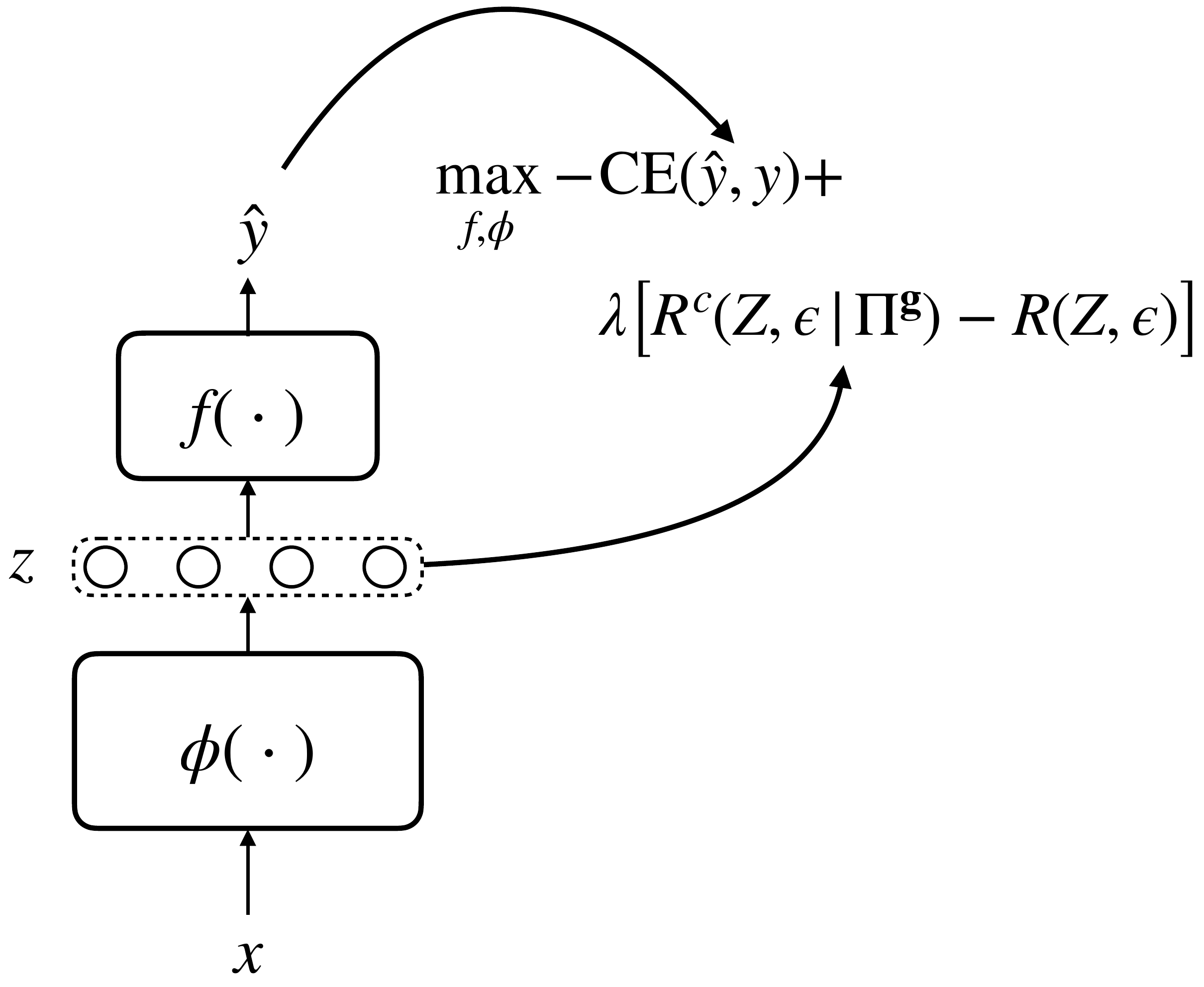}
    \caption{\small Constrained debiasing setup using {\X}. Representation $z$ retrieved from the feature map $\phi$ is used to predict the target label and control the rate-distortion objective function.}
    \label{fig:constrained}
    \vspace{-13pt}
\end{figure}

Our constrained debiasing setup is shown in Figure~\ref{fig:constrained}, where representations are retrieved from a feature map $\phi$ followed by a target task classifier $f$. In this setup, we achieve (a) by training $f$ to predict the target class $\hat{y} = f(z)$ and minimize the cross-entropy loss $\mathrm{CE}(\hat{y}, y)$, where $y$ is the ground-truth target label. For (b), we need to ensure that representations from different protected classes are similar and overlap in the representation space.  This is achieved by \textit{maximizing} the rate $R^c(Z, \epsilon|\Pi^{\mathbf{g}})$ while \textit{minimizing} $R(Z, \epsilon)$. Maximizing $R^c(Z, \epsilon|\Pi^{\mathbf{g}})$ ensures samples in the same protected class are dissimilar and have large intra-class variance. {However, simply increasing intra-class variance does not guarantee the overlap of different protected class representations -- as the overall feature space can expand and representations can still be discriminative w.r.t $\mathbf g$. Therefore, we also minimize} $R(Z, \epsilon)$ ensuring a lower number of bits are required to encode all representations $Z$, thereby making the representation space compact. This process is illustrated visually in Figure~\ref{fig:constrained-viz}. The \textcolor{blue}{\textbf{blue}} and \textcolor{red}{\textbf{red}} circles correspond to representations 
from two protected classes. The \textcolor{gray}{\textbf{gray}} arrows are induced by the term $R^c(Z, \epsilon|\Pi^{\mathbf{g}})$ that encourages the representations to be dissimilar to samples in the same protected class. The \textcolor{green}{\textbf{green}} arrows 
induced by $R(Z, \epsilon)$ try to make the representation space more compact. To achieve this objective, {\X} adds a {rate-distortion based} regularization constraint {to} the target classification loss. 
{Overall, {\X} achieves (a) and (b) simultaneously by \textit{maximizing} the following objective function:} 

\begin{equation}
    \begin{aligned}
    J_{\mathrm c}(Z, Y, \Pi^{\mathbf g}) &= -\mathrm{CE}(\hat{y}, y) \\&+ \lambda \big[R^c(Z, \epsilon|\Pi^{\mathbf g}) - R(Z, \epsilon)\big]
    \end{aligned}
    \label{eqn:constrained-loss}
\end{equation}


\noindent where $\hat{y}$ 
is the target prediction label, $y$ is the ground-truth label and 
$\lambda$ is a hyperparameter.\footnote{Note, we cannot use the same regularization term (Equation~\ref{eqn:constrained-loss}) for unconstrained debiasing, as minimizing $R(Z, \epsilon)$ without the supervision of target loss $\mathrm{CE}(\hat{y}, y)$ results in all representations converging to a compact space,  thereby losing most of the information. We have verified this experimentally (more details in Appendix~\ref{sec:ablation}).
} \SBB{We select the hyperparameters using grid search and discuss the hyperparameter sensitivity of {\X} in Section~\ref{sec:lambda}. We follow a similar routine to obtain debiased representations in the constrained setup as shown in Algorithm~\ref{alg:unconstrained}.}

\section{Experimental Setup}
In this section, we discuss the datasets, experimental setup, and metrics used for evaluating {\X}. \SBB{The implementation of {\X} is publicly available \href{https://github.com/brcsomnath/FaRM}{https://github.com/brcsomnath/FaRM}.}

\subsection{Datasets} 
\label{sec:dataset}
We evaluate {\X} using several datasets.
Among these, the {\dial} and Biographies datasets are used for {evaluating both  constrained and unconstrained debiasing}. 
{\pan} and GloVe embeddings are used only for constrained and unconstrained debiasing respectively. We use the same train-test split as prior works for all datasets.

\noindent(a) \textbf{{\dial}} \cite{blodgett2016demographic} is a Twitter-based sentiment classification dataset. Each tweet
is associated with sentiment and mention labels ({treated as the \textit{target attribute} in constrained evaluation}) and ``race'' information ({\textit{protected attribute}}) of the author. The sentiment labels are ``happy'' or ``sad'' and the race categories are ``African-American English'' (AAE) or ``Standard American English'' (SAE). 

\noindent(b) \textbf{Biography classification} dataset~\cite{de2019bias} contains biographies that are associated with a profession (\textit{target attribute}) and gender label (\textit{protected attribute}). There are 28 distinct profession categories and 2 gender classes. 



 \noindent(c) \textbf{{\pan}} \cite{rangel2016overview} is also a Tweet-classification dataset where each Tweet is annotated with the author's {age} and {gender} information, both of which are {binary protected attributes}. The {target task} is {mention detection}. 


\noindent(d) \textbf{GloVe embeddings}: We follow the setup of \citet{ravfogel2020null} to debias the most common 150,000 Glove word embeddings \cite{zhao2018learning}. For training, we use the 7500 most male-biased, female-biased, and neutral words (determined by the magnitude of the word vector's projection onto the 
gender direction, which is the largest principal component of the space of vectors formed using the difference gendered word vector pairs). 

\subsection{Implementation details}
We use a mutli-layer neural network with ReLU non-linearity as our feature map $\phi$ in the unconstrained setup. This setup is optimized using stochastic gradient descent with a learning rate of 0.001 and momentum of 0.9. For constrained debiasing, we used BERT\textsubscript{base} as $\phi$, and a 2-layer neural network as $f$. Constrained setup is optimized using AdamW~\cite{loshchilov2017decoupled} optimizer with a learning rate of 2$\times10^{-5}$. We set $\lambda=0.01$ for all experiments. Hyperparameters were tuned on the development set of the respective datasets. Our models were trained on a single Nvidia Quadro RTX 5000 GPU.

\subsection{Probing Metrics}
Following \cite{elazar2018adversarial, ravfogel2020null, basu-roy-chowdhury-etal-2021-adversarial}, we evaluate the quality of our debiasing by probing the learned representations 
for the protected attribute $\mathbf g$ and target attribute $\mathbf y$. {In our experiments, we probe all representations using a non-linear classifier. {We use an MLP Classifier from the scikit-learn library \cite{pedregosa2011scikit}.}} We report the Accuracy and Minimum Description Length (MDL)~\cite{voita2020information} {for predicting $\mathbf g$ and $\mathbf y$}. 
A {large} MDL signifies that more effort is needed by a probing network to achieve a certain performance. 
Hence,  we expect debiased representations to have a \textit{large} MDL for protected attribute $\mathbf g$ and a \textit{small} MDL for predicting target attribute $\mathbf y$. Also, we expect a \textit{high} accuracy for $\mathbf y$ and \textit{low} accuracy for $\mathbf g$. 

\subsection{Group Fairness Metrics} 

\noindent\textbf{TPR-GAP.} Based on the notion of \textit{equalized odds}, \citet{de2019bias} introduced TPR-GAP -- which measures the true positive rate (TPR) difference of a classifier between two protected groups. 
TPR-GAP for a target attribute label $y$ is:

\begin{equation*}
    \small
    \begin{aligned}
        \mathrm{TPR}_{\mathbf{g},y} &=p(\hat{\mathbf y} = y|\mathbf{g} = g, \mathbf{y} = y)\\
        \mathrm{Gap}_{\mathbf{g},y} &= \mathrm{TPR}_{g,y} - \mathrm{TPR}_{\bar{g},y}
    \end{aligned}
\end{equation*}

\noindent where $\mathbf y$ is the target attribute, $\mathbf g$ is a binary protected attribute with possible values
$g, \bar{g}$, and $\hat{\mathbf y}$ denotes the predicted target attribute. 
 \citet{biasbios2} proposed a single bias score 
 for the classifier called $\mathrm{Gap}_\mathbf{g}^{\mathrm{RMS}}$, which is defined as:
\begin{equation}
\small
\mathrm{Gap}_\mathbf{g}^{\mathrm{RMS}} = \sqrt{\frac{1}{\lvert \mathcal{Y}\rvert} \sum_{y \in \mathcal{Y}} (\mathrm{Gap}_{\mathbf{g},y})^2}
\end{equation}

\noindent where is $\mathcal Y$ is the set of target attribute 
labels.

\noindent\textbf{Demographic Parity (DP).} DP measures the difference in prediction w.r.t to protected attribute $\mathbf g$. 
\begin{equation*}
    \small
    \mathrm{DP} = \sum\limits_{y \in \mathcal{Y}}\lvert p(\hat{\mathbf y} = y|\mathbf{g} = g) -  p(\hat{\mathbf y} = y|\mathbf{g} = \bar{g})\rvert
    \label{eqn:DP}
\end{equation*}

\noindent where $g, \bar{g}$ are possible values of the binary protected attribute $\mathbf g$ and $\mathcal{Y}$ is the set of possible target attribute labels. 

\label{sec:fair-metrics}
\citet{bickel1975sex} illustrated that notions of demographic parity and equalized odds can strongly differ in a real-world scenario. For representation learning, \citet{zhao2019inherent} demonstrated an inherent tradeoff between the utility and fairness of representations. TPR-GAP {described above} is not a good indicator of fairness if $\mathbf y$ and $\mathbf g$ are correlated, as debiasing would lead to a drop in target task performance as well. 
{For our experiments, we compare models using both metrics for completeness. However, like prior works, in some cases we observe conflicting results due to the tradeoff.}

\begin{table}[t!]
    \centering
    \resizebox{0.38\textwidth}{!}{
		\begin{tabular}{ l c c c} 
			\toprule[1pt]
			{Method} & Accuracy ($\downarrow$) & MDL ($\uparrow$) & Rank ($\uparrow$) \TBstrut\\
			\midrule[1pt]
            GloVe & {100.0} & {0.1} & {300} \\
            INLP & 86.3 & 8.6 & 210 \\
            {\X} & \textbf{53.9} & \textbf{24.6} & \textbf{247} \\
			\bottomrule[1pt]
	\end{tabular}
}
    \caption{\small Debiasing performance on Glove word embeddings. 
    {\X} significantly outperforms INLP~\cite{ravfogel2020null} in guarding gender information. Best debiasing results are in \textbf{bold}.}
    \label{tab:results-glove}
    \vspace{-15pt}
\end{table}

\section{Results: Unconstrained Debiasing}
We evaluate {\X} for unconstrained debiasing in three different setups: 
word embedding debiasing, {and debiasing text representations for } 
biographies  and  sentiment classification.
For the classification tasks, we retrieve text representations from a pre-trained encoder, debias them using {\X} (without taking the target task into account) and evaluate the debiased representations by probing for $\mathbf y$ and $\mathbf g$. \SBB{In all settings, we train the feature encoder $\phi$, and evaluate the retrieved representations $Z_{\text{debiased}} = \phi(X)$.}
All tables mention the expected trend of a metric using $\uparrow$ (higher) or $\downarrow$ (lower).

\subsection{Word Embedding Debiasing}
We revisit the problem of debiasing gender information from word embeddings introduced by \citet{bolukbasi2016man}.

\noindent \textbf{Setup.} We debias gender information from GloVe embeddings using a 4-layer neural network with ReLU non-linearity as the feature map $\phi(x)$. {We discuss the choice of the feature map $\phi$ in Section~\ref{subsec:limitations}.}

\noindent \textbf{Results.} Table~\ref{tab:results-glove} presents the result of debiasing word embeddings for baseline INLP~\cite{ravfogel2020null} and {\X}. 
{We observe that when compared with INLP, {\X} reduces the accuracy of the network by an absolute margin of 32.4\% and achieves a steep increase in MDL. {\X} is able to guard the protected attribute against an attack by a non-linear probing network  (near-random accuracy). \SBB{We also report the rank of the resulting word embedding matrix. The information content of a matrix is captured by its rank (maximal number of linearly independent columns). An increase in rank of the resultant embedding matrix indicates that {\X} is able to retain more information in the representations, in general, compared to INLP.} }

\begin{figure}[t!]
    \centering
    \subfloat[][\small GloVe]{
        \includegraphics[width=0.225\textwidth, keepaspectratio]{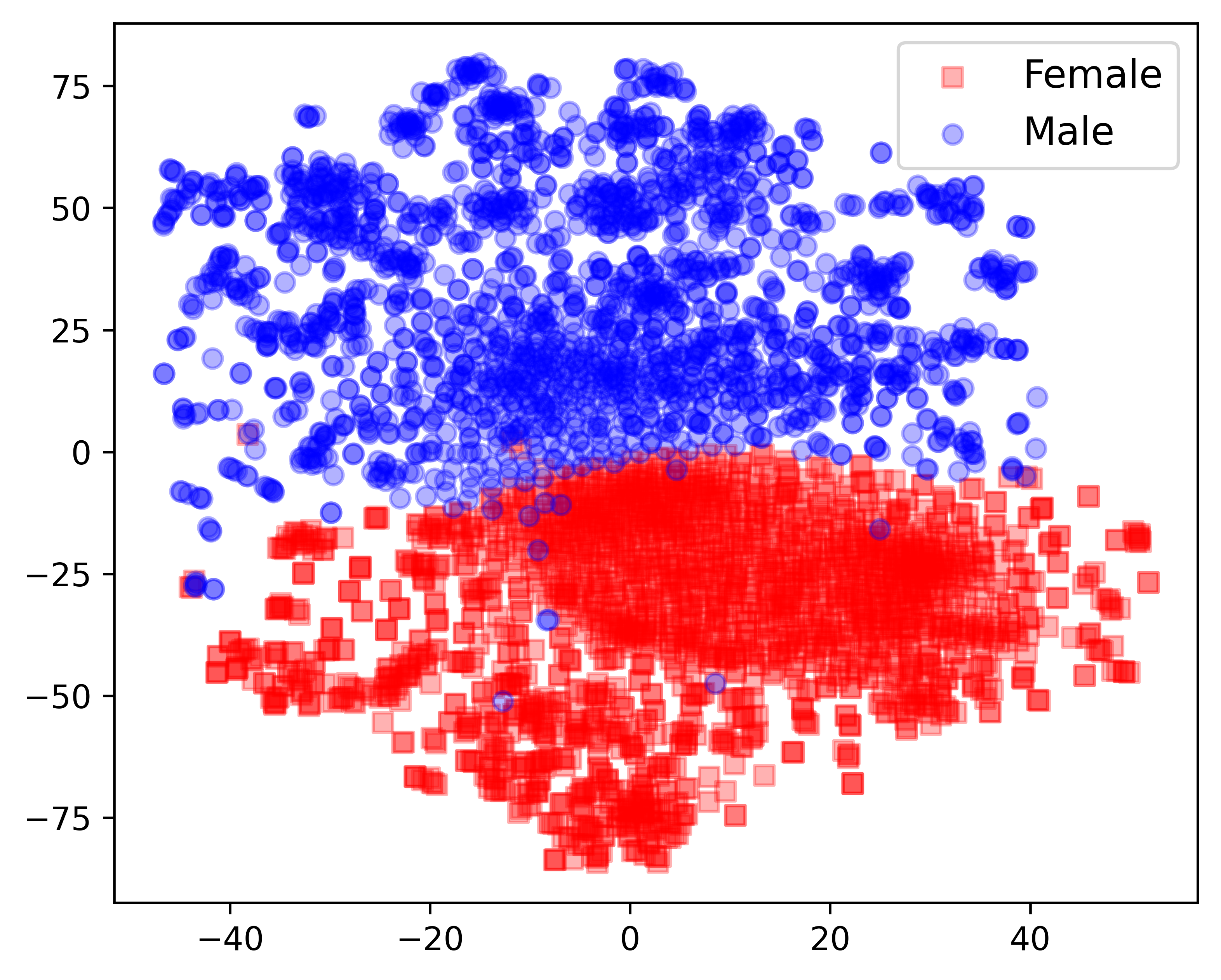}{ 
            \label{subfig:original}}
    }
    \subfloat[][\small Debiased]{
        \includegraphics[width=0.225\textwidth, keepaspectratio]{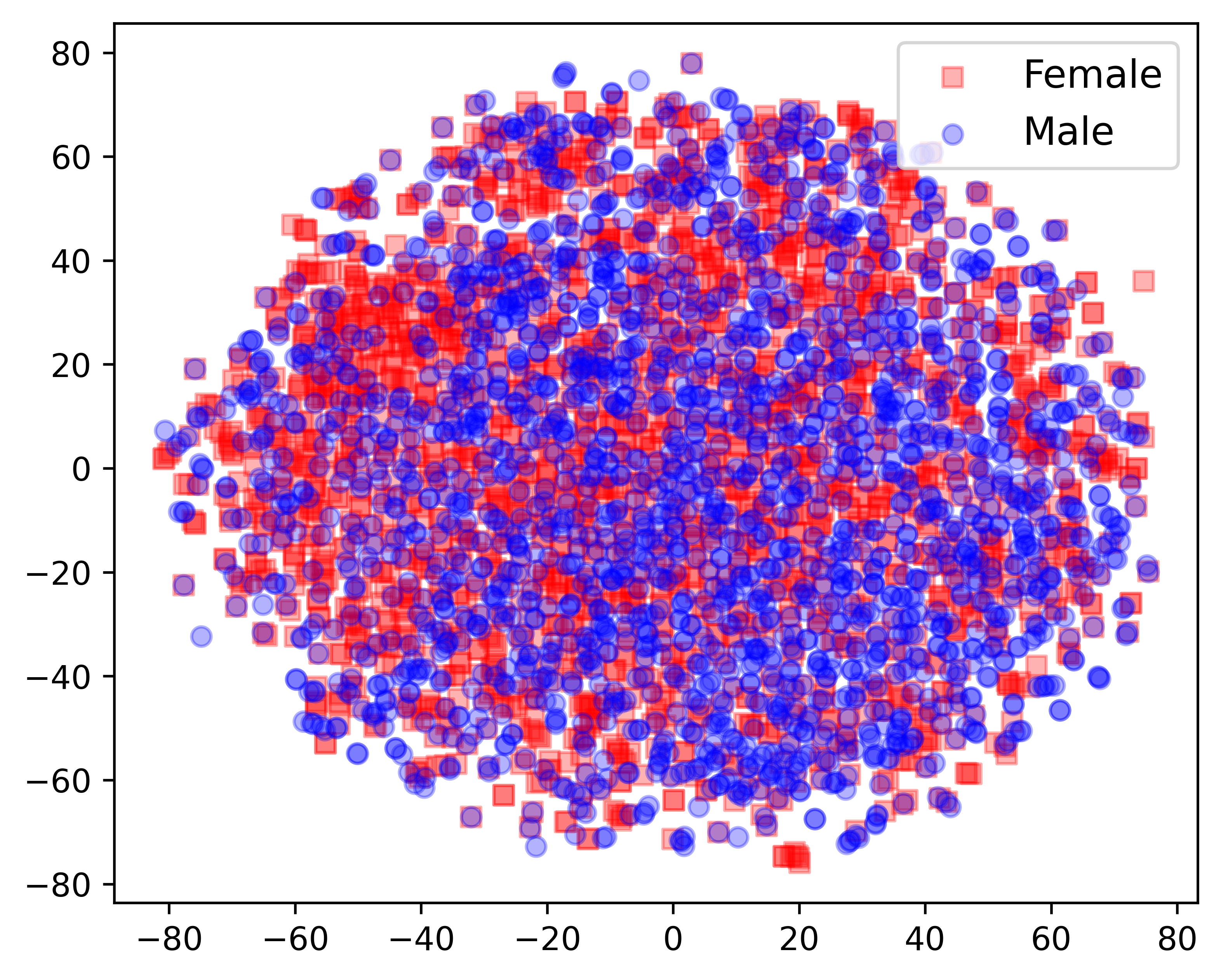}{ 
            \label{subfig:debiased}}
    }
    \caption{\small Projections of Glove embeddings before (left) and after (right) debiasing. Intial female and male biased representations are shown in \textcolor{red}{\textbf{red}} and \textcolor{blue}{\textbf{blue}} respectively.}
    \label{fig:tsne-glove}
    \vspace{-12pt}
\end{figure}

\noindent\textbf{Visualization.}  We visualize the t-SNE~\cite{van2008visualizing} projections of Glove embeddings before and after debiasing in Figures~\ref{subfig:original} and~\ref{subfig:debiased} respectively. Female and male-biased word vectors 
are represented by \textcolor{red}{\textbf{red}} and \textcolor{blue}{\textbf{blue}} dots respectively.  The figures clearly demonstrate that the gendered vectors are not separable after debiasing. In order to quantify the improvement, we perform $k$-means clustering with $K=2$ (one for each gender label). We compute the V-measure \cite{v-measure} -- a measure to quantify the overlap between clusters. V-measure in the original space drops from 99.9\% to 0.006\% using {\X} (compared to 0.31\% using INLP). {This indicates that debiased representations from {\X} are more difficult to disentangle}. We further analyze the quality of the debiased word embeddings in Section~\ref{sec:probing-we}.


\subsection{Biography Classification}
\label{sec:unconstrained-bios}
{Next, we evaluate {\X} by debiasing text representations in an unconstrained setup and using the representations for fair biography classification.}

\noindent \textbf{Setup.} We obtain the text representations using {two methods:} FastText \cite{fasttext} and BERT \cite{bert}. For FastText, we sum the individual token representations in each biography. For BERT, by retrieving the final layer hidden representation above the \texttt{[CLS]} token from a pre-trained BERT\textsubscript{base} model. We choose the feature map $\phi(x)$ as a 4-layer 
neural network with ReLU non-linearity.

\begin{table}[t!]
    \centering
    \resizebox{0.41\textwidth}{!}{
		\begin{tabular}{c l c c} 
			\toprule[1pt]
			Metric & Method & FastText & BERT \TBstrut\\
			\midrule[1pt]
			\multirow{3}{*}{\shortstack[c]{Profession \\ Acc.}} \multirow{3}{*}{($\uparrow$)} & Original & 79.9 & 80.9\\
			& INLP & \textbf{76.3} & \textbf{77.8}\\
			& {\X} & 54.8 &  55.8\\
			\midrule[1pt]
			\multirow{3}{*}{\shortstack[c]{Gender \\Acc.}} \multirow{3}{*}{($\downarrow$)} & Original & 98.9 & 99.6\\
			& INLP & 67.4 & 94.9\\
			& {\X} & \textbf{57.6} &  \textbf{55.6}\\
            \midrule[1pt]
            \multirow{3}{*}{DP ($\downarrow$)} & Original & 1.65 & 1.68\\
			& INLP & 1.51 & 1.50\\
			& {\X} & \textbf{0.12} & \textbf{0.14} \\
			\midrule[1pt]
			\multirow{3}{*}{$\mathrm{Gap}_\mathbf{g}^{\mathrm{RMS}}$ ($\downarrow$)} & Original & 0.185 & 0.171\\
			& INLP & 0.089 & 0.096\\
			& {\X} & \textbf{0.006} & \textbf{0.079} \\
			\bottomrule[1pt]
	\end{tabular}
}
    \caption{\small Evaluation results of {\X} on Biographies dataset. Compared to INLP~\cite{ravfogel2020null}, representations from {\X} leak significantly less gender information and achieve better fairness scores. 
    }
    \label{tab:results-bios}
    \vspace{-12pt}
\end{table}

\noindent \textbf{Results.} Table~\ref{tab:results-bios} presents the unconstrained debiasing results of {\X} on this dataset. 
`Original' in the table refers to the pre-trained embeddings from BERT\textsubscript{base} or FastText.
We observe that {\X} significantly outperforms INLP in fairness metrics -- DP (improvement of 92\% for FastText and 91\% for BERT) and $\mathrm{Gap}_\mathbf{g}^{\mathrm{RMS}}$ (improvement of 93\% for FastText and 18\% for BERT). 
We observe that {\X} achieves near-random gender classification performance (majority baseline: 53.9\%) against a non-linear probing attack.  {\X} improves upon INLP's gender leakage by an absolute margin of 9.8\% and 39.4\% for FastText and BERT respectively.  However, we observe a substantial drop in the accuracy for identifying professions (target attribute) using the debiased embeddings.\footnote{Majority baseline for profession classification  $\approx$29\%.} \SBB{This is possibly because in this dataset, gender is highly correlated with the profession and removing gender information results in loss of profession information. \citet{zhao2019inherent} identified this phenomenon by noting the tradeoff between learning fair representations and performing well on target task, when protected and target attributes are correlated. The results in this setup (Table~\ref{tab:results-bios}) demonstrate this phenomenon. In unconstrained debiasing, we remove information about protected attribute from the representations without taking into account the target task. As a result target task performance suffers from more debiasing.}\footnote{In our experiments, we found profession  accuracy to be high with a shallow feature map or training for earlier epochs, but the gender leakage was significant in these scenarios.} This calls for constrained debiasing for such datasets. In Section~\ref{sec:constrained-results}, we show that {\X} is able to retain target performance while debiasing for this dataset in the constrained setup. 

\begin{table}[t!]
    \centering
    \resizebox{0.48\textwidth}{!}{
		\begin{tabular}{c l c c c c} 
			\toprule[1pt]
			\multirow{2}{*}{Metric} & \multirow{2}{*}{Method}  & \multicolumn{4}{c}{Split} \TBstrut\\
			 & & 50\% & 60\% & 70\% & 80\% \Bstrut\\
			\midrule[1pt]
			\multirow{3}{*}{\shortstack[c]{Sentiment \\ Acc.}} \multirow{3}{*}{($\uparrow$)} & Original & 75.5 & 75.5 & 74.4 & 71.9\\
			& INLP & \textbf{75.1} & {73.1} & \textbf{69.2} & \textbf{64.5}\\
			& {\X} & 74.8 & \textbf{73.2} & {67.3} & 63.5\\
			\midrule[1pt]
			\multirow{3}{*}{\shortstack[c]{Race \\Acc. }} \multirow{3}{*}{($\downarrow$)} & Original & 87.7 & 87.8 & 87.3 & 87.4\\
			& INLP & 69.5 & 82.2 & 80.3 & 69.9\\
			& {\X} & \textbf{54.2} & \textbf{69.9} & \textbf{69.0} & \textbf{52.1}\\
            \midrule[1pt]
            \multirow{3}{*}{DP ($\downarrow$)} & Original & 0.26 & 0.44 & 0.63 & 0.81\\
			& INLP & 0.16 & 0.33 & 0.30 & 0.28\\
			& {\X} & \textbf{0.09} & \textbf{0.10} & \textbf{0.17} & \textbf{0.22} \\
			\midrule[1pt]
			\multirow{3}{*}{$\mathrm{Gap}_\mathbf{g}^{\mathrm{RMS}}$ ($\downarrow$)} & Original & 0.15 & 0.24 & 0.33 & 0.41\\
			& INLP & 0.12 & 0.18 & 0.16 & 0.16\\
			& {\X} & \textbf{0.09} & \textbf{0.10} & \textbf{0.12} & \textbf{0.14} \\
			\bottomrule[1pt]
	\end{tabular}
}
    \caption{\small Evaluation results of unconstrained debiasing on {\dial} dataset. We report the performance of the DeepMoji  (Original), INLP~\cite{ravfogel2020null} and {\X} representations. We observe that {\X} achieves the best fairness scores in all setups, while maintaining similar performance on sentiment classification task. 
    }
\label{tab:dial}
    \vspace{-13pt}
\end{table}

\subsection{Controlled Sentiment Classification}
Lastly,  for the {\dial} dataset, we perform unconstrained debiasing in a controlled setting. 

\begin{table*}[t!]
    \centering
    \resizebox{\textwidth}{!}{
	\begin{tabular}{l|c c | c c | c c| c c | c c | c c}
	\toprule[1pt]
         \multirow{3}{*}{Method} & \multicolumn{12}{c}{\textsc{Dial}}\\
         & \multicolumn{2}{c}{Sentiment ($\mathbf y$)} & \multicolumn{2}{c}{Race ($\mathbf g$)}  & \multicolumn{2}{c|}{Fairness}  & \multicolumn{2}{c}{Mention ($\mathbf y$)} & \multicolumn{2}{c}{Race ($\mathbf g$)} & \multicolumn{2}{c}{Fairness} \\
         & F1$\uparrow$ & MDL$\downarrow$ & $\Delta$F1$\downarrow$ & MDL$\uparrow$ & DP$\downarrow$ & $\mathrm{Gap}_{\mathbf g}^{\mathrm{RMS}}$ $\downarrow$ & F1$\uparrow$ & MDL$\downarrow$ & $\Delta$F1$\downarrow$ & MDL$\uparrow$ & DP$\downarrow$ & $\mathrm{Gap}_{\mathbf g}^{\mathrm{RMS}}$ $\downarrow$\\
     \midrule[1pt]
         BERT\textsubscript{base} (pre-trained) & 63.9 & 300.7 & 10.9 & 242.6 & 0.41 & 0.20 & 66.1 & 290.1 & 24.6 & 258.8 & 0.20 & 0.10  \\
         BERT\textsubscript{base} (fine-tuned) & \textbf{76.9} & 99.0 & 18.4 & 176.2 & 0.30 & 0.14 & \textbf{81.7} & 49.1 & 28.7 & 199.2 &  {0.06} &  {0.03} \\
         AdS & 72.9 & 56.9 & 5.2 & 290.6  & 0.43 & 0.21 & {81.1} & 7.6 & 21.7 & 270.3  &  {0.06} &  {0.03}\\
         {\X} & {73.2} & \textbf{17.9} & \textbf{0.2} & \textbf{296.5} & \textbf{0.26} & \textbf{0.14} & 78.8 & \textbf{3.1} & \textbf{0.3} & \textbf{324.8} &  {0.06} &  {0.03}\\
    \bottomrule[0.5pt]
    \end{tabular}
}
\resizebox{\textwidth}{!}{
	\begin{tabular}{l|c c | c c | c c | c c | c c| c c}
	\toprule[0.5pt]
         \multirow{3}{*}{Method} & \multicolumn{12}{c}{\textsc{Pan16}} \\
         & \multicolumn{2}{c}{Mention ($\mathbf y$)} & \multicolumn{2}{c}{Gender ($\mathbf g$)} & \multicolumn{2}{c|}{Fairness} & \multicolumn{2}{c}{Mention ($\mathbf y$)} & \multicolumn{2}{c}{Age ($\mathbf g$)} & \multicolumn{2}{c}{Fairness}\\
         & F1$\uparrow$ & MDL$\downarrow$  & $\Delta$F1$\downarrow$ & MDL$\uparrow$ & DP$\downarrow$ & $\mathrm{Gap}_{\mathbf g}^{\mathrm{RMS}}\downarrow$ & F1$\uparrow$ & MDL$\downarrow$ & $\Delta$F1$\downarrow$ & MDL$\uparrow$  & DP$\downarrow$ & $\mathrm{Gap}_{\mathbf g}^{\mathrm{RMS}}\downarrow$\\
     \midrule[1pt]
         BERT\textsubscript{base} (pre-trained) & 72.3 & 259.7 & 7.4 & 300.5 & 0.11 & 0.056 & 72.8 & 262.6  & 6.1 & {302.0} & 0.14 & 0.078\\
         BERT\textsubscript{base} (fine-tuned) & \textbf{89.7} & {4.0} & 15.1 & 267.6 &  {0.04} &  {0.007} & \textbf{89.3} & 4.8 & 7.4 & 295.4 & 0.04 & 0.006 \\
         AdS &  {89.7} & 7.6 & {4.9} & \textbf{313.9} &  {0.04} &  {0.007} & {89.2} & 6.0 & {1.1} & \textbf{315.1}  & {0.04} & \textbf{0.004}\\
         {\X} & 88.7 & \textbf{1.7} & \textbf{0.0} & 312.4 & {0.04} &  {0.007} & 88.6 & \textbf{0.8} & \textbf{0.0} & 312.6  & \textbf{0.03} & {0.008}\\
    \bottomrule[0.5pt]
    \end{tabular}
}
\resizebox{0.63\textwidth}{!}{
	\begin{tabular}{l|c c | c c | c c}
	\toprule[0.5pt]
         \multirow{3}{*}{Method} & \multicolumn{6}{c}{\textsc{Biographies}} \\
          & \multicolumn{2}{c}{Profession ($\mathbf y$)} & \multicolumn{2}{c}{Gender ($\mathbf g$)} & \multicolumn{2}{c}{Fairness}\\
         & F1$\uparrow$ & MDL$\downarrow$ & $\Delta$F1$\downarrow$ & MDL$\uparrow$  & DP$\downarrow$ & $\mathrm{Gap}_{\mathbf g}^{\mathrm{RMS}}$ $\downarrow$\\
     \midrule[1pt]
         BERT\textsubscript{base} (pre-trained) & 74.3 & 499.9 &  45.2 & 27.6 & {0.43} & 0.169\\
         BERT\textsubscript{base} (fine-tuned) &  99.9 & \textbf{2.2} & 8.3 & 448.9 & 0.46 & \textbf{0.001} \\
         AdS  & 99.9 & {3.3} & \textbf{3.1} & {449.5}  & 0.45 & 0.003\\
         {\X} & {99.9} & 7.6 & 7.4 & \textbf{460.3}  & \textbf{0.42} & {0.002}\\
    \bottomrule[1pt]
    \end{tabular}
    }

    \caption{\small Evaluation results for constrained debiasing on {\dial}, {\pan} and Biographies. For {\dial} and {\pan}, we evaluate the approaches for two different configurations of target and proteccted variables, and report the performances in each setting.
    {\X} outperforms AdS \cite{basu-roy-chowdhury-etal-2021-adversarial} in DP metric in all setups, while achieving comparable target task performance.}
    \label{tab:constrained-results}
    \vspace{-10pt}
\end{table*}

\noindent \textbf{Setup.} Following the setup of \citet{barrett-etal-2019-adversarial, ravfogel2020null}, we control the proportion of protected attributes within a target task class. For example, if target class split=80\% that means ``happy'' sentiment (target) class contains 80\% AAE / 20\% SAE, while the ``sad” class contains 20\% AAE / 80\% SAE (AAE and SAE are protected class labels mentioned in Section~\ref{sec:dataset}).
We train DeepMoji \cite{felbo2017using} followed by a 1-layer MLP for sentiment classification. We retrieve representations from the DeepMoji encoder and debias them using {\X}. For debiasing,  we choose the feature map $\phi(x)$ to be a 7-layer neural network with ReLU non-linearity.  After debiasing, we train a non-linear MLP to investigate the quality of debiasing. We evaluate the debiasing performance of {\X} in various stages of label imbalance.

\noindent \textbf{Results.} {The results of this experiment are reported in Table~\ref{tab:dial}.}
We see that {\X} is able to achieve the best fairness scores -- an improvement in $\mathrm{Gap}_\mathbf{g}^{\mathrm{RMS}}$ ($\geq$ 12.5\%) and DP ($\geq$ 21\%) across all setups. Considering the accuracy of identifying the protected attribute -- race -- we can see that {\X} significantly reduces leakage of race information  by an absolute margin of 11\%-17\% across different target class splits.   {\X} also achieves similar performance to INLP in sentiment (target attribute) classification. We observe that the fairness score for {\X} deteriorates with an increasing correlation between the protected attribute and the target attribute. In cases where the target and the protected attributes are highly correlated (split=70\% and 80\%), we observe a low sentiment classification accuracy (for both INLP and {\X}) compared to the original classifier. This is similar to the observation made for the Biographies dataset and shows that it is difficult to debias {information about protected attribute}  while retaining overall information about the target task when the protected attribute is highly correlated with the target attribute. In the constrained setup, we observe {\X} is able to retain target performance (Section~\ref{sec:constrained-results}). 

\section{Results: Constrained Debiasing}
\label{sec:constrained-results}
In this section, we present the results of constrained debiasing using {\X}. 
 For all experiments, we use a BERT\textsubscript{base} model as $\phi$ and a 2-layer neural network with ReLU non-linearity as $f$ (Figure~\ref{fig:constrained}). 
\subsection{Single Attribute Debiasing} 
In this setup, we focus on debiasing a single protected attribute $\mathbf g$ while retaining information about the target attribute $\mathbf y$.

\noindent \textbf{Setup.}   We conduct experiments on 3 datasets: {\dial}~\cite{blodgett2016demographic}, {\pan}~\cite{rangel2016overview}, and Biographies~\cite{de2019bias}. We experiment with different target and protected attribute configurations in {\dial} ($\mathbf y$: Sentiment/Mention, $\mathbf g$: Race) and {\pan} ($\mathbf y$: Mention, $\mathbf g$: Gender/Age). For Biographies, we use the same setup as described in Section~\ref{sec:unconstrained-bios}. 
For the protected attribute $\mathbf g$, we report $\Delta$F1 -- the difference between F1-score and the majority baseline. 
We also report fairness metrics: $\mathrm{Gap}_\mathbf{g}^{\mathrm{RMS}}$ and Demographic Parity 
(DP) of the learned classifier. We compare {\X} with the state-of-the-art AdS~\cite{basu-roy-chowdhury-etal-2021-adversarial},  BERT\textsubscript{base} sequence classifier, and pre-trained BERT\textsubscript{base} representations.

\noindent \textbf{Results.} Table~\ref{tab:constrained-results} presents the {results of this experiment.} 
{We observe that in general, {\X} achieves good fairness performance while maintaining target performance. In particular, it achieves the best DP scores across all setups. In \pan, {\X} achieves perfect fairness in terms of protected attribute probing accuracy $\Delta\mathrm{F1} = 0$ with comparable performance to AdS in terms of MDL of $\mathbf g$. In the Biographies dataset,  the task accuracy of {\X} is the same as AdS but {\X} outperforms AdS in fairness metrics. We also observe that for this dataset, some baselines performed very well on one (but not both) of the two fairness metrics which can be attributed to the inherent trade-off between them (see Section~\ref{sec:fair-metrics}).
However, {\X} achieves a good balance between the two metrics.} Overall, this shows that {\X} is able to robustly remove sensitive information about the protected attribute while achieving good target task performance.


\subsection{Multiple Attribute Debiasing}
In this setup, we focus on debiasing multiple protected attributes $\mathbf{g}_i$ simultaneously, while retaining information about target {attribute} $\mathbf y$. We evaluate {\X} on the {\pan} dataset with $\mathbf y$ as Mention, $\mathbf{g}_1$ as Gender, and $\mathbf{g}_2$ as Age. 
\citet{subramanian2021evaluating} showed that debiasing a categorical attribute can still reveal information about {intersectional groups} (e.g. if {age} (young/old) and {gender} (male/female) are two categorical protected attributes, then (age=old, gender=male) is an intersectional group). We report the $\Delta$F1/MDL scores for probing intersectional groups. 

\begin{table*}[t!]
	\centering
	\resizebox{\textwidth}{!}{
        \begin{tabular}{l| c c| c c|c c |c c| c c | c c}
	\toprule[1pt]
	 \multirow{3}{*}{\textsc{Setup}} & \multicolumn{12}{c}{\textsc{Pan16}}\TBstrut\\
	 & \multicolumn{2}{c|}{Mention ($y$)} & \multicolumn{2}{c}{Age ($\mathbf{g}_1$)} & \multicolumn{2}{c|}{Fairness ($\mathbf{g}_1$)} & \multicolumn{2}{c}{Gender ($\mathbf{g}_2$)} & \multicolumn{2}{c|}{Fairness ($\mathbf{g}_2$)} & \multicolumn{2}{c}{Inter. Groups ($\mathbf{g}_1,\mathbf{g}_2$)}\\
	 & F1$\uparrow$ & MDL$\downarrow$ & $\Delta$F1$\downarrow$ & MDL$\uparrow$ & DP$\downarrow$ & $\mathrm{Gap}_{\mathbf g}^{\mathrm{RMS}}\downarrow$ & $\Delta$F1$\downarrow$ & MDL$\uparrow$ & DP$\downarrow$ & $\mathrm{Gap}_{\mathbf g}^{\mathrm{RMS}}\downarrow$ & $\Delta$F1$\downarrow$ & MDL$\uparrow$ \Bstrut\\
	\midrule[0.5pt]
	BERT\textsubscript{base} (fine-tuned) & 88.6 & 6.8 & 14.9 & 196.4 & 0.06 & 0.009 & 16.5 & 192.0 & 0.04 & 0.014 & 20.7 & 117.2 \Tstrut\\
	\textsc{AdS} & \textbf{88.6} & \textbf{5.5} & {2.2} & {231.5} & 0.05 & 0.006 & {1.6} & {230.9} & 0.04 & 0.017 & 9.1 & 118.5\\
	{\X} ($N$-partition) & 87.0 & 13.4 &   {0.0} & 234.3 & \textbf{0.03} & \textbf{0.003} &   {0.0} &   {234.2} & 0.06 & 0.025 & 0.7 & \textbf{468.0}\\
	{\X} (1-partition) & 86.4 & 15.6 &   {0.0} & \textbf{234.6} & 0.05 & 0.006 &   {0.0} &   {234.2} & \textbf{0.02} & \textbf{0.009}  & \textbf{0.0} & 467.7 \Bstrut\\
	\bottomrule[1pt]
\end{tabular}
	}
	\vspace{-5pt}
	\caption{\small Evaluation results for debiasing multiple protected attributes using {\X}. Both configurations of {\X}  outperform AdS \cite{basu-roy-chowdhury-etal-2021-adversarial} in guarding protected attribute and intersectional group biases.}
	\label{tab:multi-attr}
    \vspace{-15pt}
\end{table*}

\noindent \textbf{Approach.} We present two variations of {\X} to remove multiple attributes simultaneously in a constrained setup. 
Assuming there are $N$ protected attributes, the variations are discussed below:\\
\noindent (a) ${N}$\textit{-partition}: In this variation, we compute a membership matrix $\Pi^{\mathbf{g}_i}$ for each protected attribute $\mathbf{g}_i$. We modify Equation~\ref{eqn:constrained-loss} as follows:

\begin{equation*}
    \begin{aligned}
    J_c(Z, \mathbf y, & \Pi^{\mathbf{g}_1}, \ldots, \Pi^{\mathbf{g}_N}) = -\mathrm{CE}(\hat{y}, y) \\&+ \lambda \sum\limits_{i=1}^N\big[R(Z, \epsilon|\Pi^{\mathbf{g}_i}) - R(Z, \epsilon)\big]
    \end{aligned}
\end{equation*}

\noindent(b) \textit{$1$-partition}: Unlike the previous setup, we can consider each protected attribute $\mathbf{g}_i$ as an independent variable and {combine them to} form a single protected attribute $\mathcal{G}$. For each input instance, {we can represent the} $i^{th}$ protected attribute {as a one-hot vector} $\mathbf{g}_{i} \in \mathbb{R}^{|\mathbf{g}_i|}$ (where $|\mathbf{g}_i|$ is the dimension of protected attribute $\mathbf{g}_i$). Then the combined vector  $\mathcal{G} \in \mathbb{R}^{(|\mathbf{g}_1| + \ldots + |\mathbf{g}_N|)}$ {can be obtained} by concatenating individual vectors $\mathbf{g}_{i}$. {Since $\mathcal{G}$ is a concatenation of multiple vectors, we normalize $\mathcal{G}$ such that all of its elements sum to $1$. Therefore each element of $\mathcal{G}$ is either $0$ or $\frac{1}{N}$}. 
We use $\mathcal{G}$ to construct the partition function $\Pi^\mathcal{G}$, which captures information about $N$ attributes simultaneously. 
Each component of $\Pi^\mathcal{G}$ satisfies: $\sum_{j=1}^N\Pi^\mathcal{G}_j = I_{n \times n}$ and $\pi_{ij} \in \{0, \frac{1}{N}\}$. The resultant objective function takes the same form as in Equation~\ref{eqn:constrained-loss} with the modified parition function $J_c({Z, Y, \Pi^\mathcal{G}})$.

\noindent\textbf{Results.} We present the results of debiasing multiple attributes in Table~\ref{tab:multi-attr}. We observe that {\X} improves upon AdS' $\Delta$F1-score of age and gender, with $N$-partition and $1$-partition setups performing equally well. The performance on the target task is comparable with AdS, although there is a slight rise in MDL. 
It is important to note that even though AdS performs decently well in preventing leakage about $\mathbf{g}_1$ and $\mathbf{g}_2$, it still leaks a significant amount of information about the intersectional groups. In both of its configurations, {\X} is able to prevent leakage of intersectional biases while considering the protected attributes independently. This shows that robustly removing information about multiple attributes helps in preventing leakage about intersectional groups as well. 

\section{Model Analysis}
In this section, we present several analysis experiments to evaluate the functioning of {\X}.

\noindent\textbf{Robustness to label corruption.} We evaluate the robustness of {\X} by randomly sub-sampling 
instances from the dataset and modifying the protected attribute label. In Figure~\ref{subfig:unconstrained}, we report the protected attribute leakage ($\Delta$F1 score) 
from the debiased word embeddings with varying fractions of training set label corruption. We observe that {\X}'s performance degrades with an increase in label corruption. This is expected {as}, at high corruption ratios, most of the protected attribute labels are wrong, resulting in poor performance. 

In the constrained setup (Figure~\ref{subfig:constrained}), we observe that {\X} is able to debias protected attribute information ($y$-axis scale in Fig.~\ref{subfig:constrained} and~\ref{subfig:unconstrained} are different)
 even at high corruption ratios. We believe this enhanced performance (compared to unconstrained setup) is due to the additional supervision in the form of target loss, which  enables {\X} to learn robust representations even with corrupted protected attribute labels.   

\begin{figure}[t!]
    \centering
    \subfloat[][\footnotesize Unconstrained]{
    \resizebox{0.245\textwidth}{!}{
    \begin{tikzpicture}[scale=0.6]
\begin{axis}[
    xlabel={\Large Corruption ratio},
    ylabel={\Large $\Delta$F1 (\%)},
    xmin=0.0, xmax=0.5,
    ymin=0, ymax=50,
    xtick={0.1, 0.2, 0.3, 0.4, 0.5},
    ytick={10, 20, 30, 40},
    ymajorgrids=true,
    grid style=dashed,
    legend pos=outer north east
]

\addplot[
    color=black,
    mark=*,
    ]
    coordinates {
    (0.5, 85.08791453371913-50)
    (0.4, 75.2949031827287-50)
    (0.3, 68.2617404851992-50)
    (0.2, 61.339862007567326-50)
    (0.1, 57.422657467171156-50)
    (0, 53.9-50)
    };
\end{axis}
\label{position-wise}
\end{tikzpicture}{\label{subfig:unconstrained}}
    }}
    \subfloat[][\footnotesize Constrained]{
    \resizebox{0.23\textwidth}{!}{
    \begin{tikzpicture}[scale=0.6]
\begin{axis}[
    xlabel={\Large Corruption ratio},
    xmin=0.0, xmax=0.5,
    ymin=0, ymax=2,
    xtick={0.1, 0.2, 0.3, 0.4, 0.5},
    ytick={0.4, 0.8, 1.2, 1.6},
    ymajorgrids=true,
    grid style=dashed,
    legend pos=outer north east
]

\addplot[
    color=black,
    mark=*,
    ]
    coordinates {
    (0.5, 63.20765327387844 - 62.266872200647036)
    (0.4, 63.91290828376921 - 63.20765327387844) 
    (0.3, 63.70384582305112 - 63.20765327387844)
    (0.2, 63.20765327387844 - 62.68830507241103)
    (0.1, 63.20765327387844 - 62.91787682516159)
    (0, 0.0)
    };
\end{axis}
\label{position-wise}
\end{tikzpicture}{\label{subfig:constrained}}
    }}
    \caption{\small Performance of {\X} with varying fraction of corrupted training set labels in (a) unconstrained and (b) constrained debiasing setups.}
    \label{fig:glove-corruption}
    \vspace{-10pt}
\end{figure}
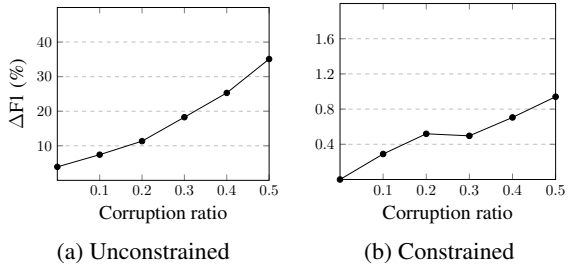

\begin{figure}[t!]
    \centering
    \resizebox{0.245\textwidth}{!}{
    \begin{tikzpicture}
\pgfplotsset{
    scale only axis,
    scaled x ticks=base 10:3,
    xmin=0.000001, xmax=10,
}

\begin{axis}[
title={\Large \textsc{Dial}},
  ymin=0, ymax=400,
  xmode=log,
  xlabel={\Large $\lambda$},
  ylabel={\Large MDL},
    ytick={100, 200, 300, 400},
    ymajorgrids=true,
    grid style=dashed,
]
\addplot[smooth,mark=*,red]
  coordinates{
    (0.000001, 204.56689186818815)
    (0.00001, 227.02626070151481)
    (0.0001, 294.08039424749774)
    (0.001, 297.41929076759544)
    (0.01, 296.5)
    (0.1, 295.9007181473673)
    (1, 301.92118543629135)
    (10, 308.88616581120885)
}; \addlegendentry{MDL ($\mathbf g$)}

\addplot[smooth,mark=square,blue]
  coordinates{
    (0.000001, 17.490821382429928)
    (0.00001, 16.275901578962575)
    (0.0001, 18.134467568219666)
    (0.001, 18.6881028159168)
    (0.01, 17.9)
    (0.1, 21.29526657338594)
    (1, 35.40473327511214)
    (10, 308.88616581120885)
}; \addlegendentry{MDL ($\mathbf y$)}

\end{axis}



\end{tikzpicture}
    }
    \resizebox{0.23\textwidth}{!}{
    \begin{tikzpicture}
\pgfplotsset{
    scale only axis,
    scaled x ticks=base 10:3,
    xmin=0.000001, xmax=10
}

\begin{axis}[
title={\Large \textsc{Pan16}},
  ymin=0, ymax=400,
  xmode=log,
  xlabel={\Large $\lambda$},
    ytick={100, 200, 300, 400},
    ymajorgrids=true,
    grid style=dashed,
]
\addplot[smooth,mark=*,red]
  coordinates{
    (0.000001, 268.35816793298346)
    (0.00001, 297.0679401872895)
    (0.0001, 309.76154867550036)
    (0.001, 311.6097012680202)
    (0.01, 312.4)
    (0.1, 312.2687289217172)
    (1, 312.4983271903098)
    (10, 313.2121609937001)
}; \addlegendentry{MDL ($g$)}

\addplot[smooth,mark=*,blue]
  coordinates{
    (0.000001, 1.4278464089336524)
    (0.00001, 1.7363602668163243)
    (0.0001, 2.3061309725638983)
    (0.001, 1.6435961669160053)
    (0.01, 1.7)
    (0.1, 5.286511115388991)
    (1, 8.216152123526658)
    (10, 313.36741730011147)
}; \addlegendentry{MDL ($y$)}

\end{axis}



\end{tikzpicture}
    }
    \caption{\small MDL of target ($\mathbf y$) and protected ($\mathbf g$) attributes with different $\lambda$ for {\dial} and {\pan} datasets.}
    \label{fig:lambda-sensitivity}
    \vspace{-17pt}
\end{figure}
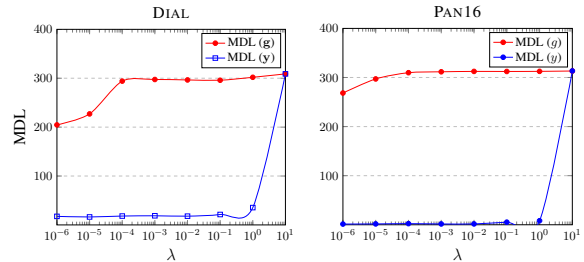

\noindent\textbf{Sensitivity to $\lambda$.}\label{sec:lambda} 
We measure the sensitivity of {\X}'s performance w.r.t $\lambda$ (Equation~\ref{eqn:constrained-loss}) in the constrained setup. In Figure~\ref{fig:lambda-sensitivity}, we show the MDL of the target attribute $\mathbf y$ (in \textcolor{blue}{\textbf{blue}}) and protected attribute $\mathbf g$ (in \textcolor{red}{\textbf{red}}) for {\dial} and {\pan} for different $\lambda$. 
We observe that when $10^{-4} \leq \lambda \leq 1$, the performance of {\X} does not change much. For $\lambda=10$, MDL for $\mathbf y$ is quite large, showcasing that the model does not converge on the target task. This is expected as the  regularization term 
(Equation~\ref{eqn:constrained-loss}) is much larger than $\mathrm{CE}(\hat{y}, y)$ term, and boosting it further with $\lambda=10$ makes it difficult for the target task loss to converge. Similarly, when $\lambda \leq 10^{-5}$, the regularization term is much smaller compared to $\mathrm{CE}(\hat{y}, y)$, there is a substantial drop in  MDL for $\mathbf g$. However, we show that {\X} achieves good performance over a broad spectrum of $\lambda$. Therefore, reproducing the desired results does not require extensive hyperparameter tuning.

\noindent\textbf{Probing Word Embeddings.}\label{sec:probing-we} A limitation of using {\X} for debiasing word embeddings is that distances in the original embedding space are not preserved. The Mazur–Ulam theorem \cite{jamison2002isometries} states that isometry for a mapping $\phi: V \rightarrow W$ is preserved only if the function $\phi$ is affine. {\X} uses a non-linear feature map $\phi(x)$. Therefore, distances cannot be preserved. 
A linear map $\phi(x)$ is also not ideal because it does not guard protected attributes against an attack by a non-linear probing network. 
We investigate the utility of debiased embeddings by performing the following experiments: 

\noindent (a) \textit{Word Similarity Evaluation}: In this experiment, 
we evaluate the debiased embeddings on the following datasets: SimLex-999 \cite{hill2015simlex}, WordSim-353 \cite{agirre2009study}, and MTurk-771 \cite{halawi2012large}. 
In Table~\ref{tab:results-sim}, we report the Spearman correlation between the gold similarity scores of word pairs and the cosine similarity scores obtained before (top row) and after (bottom row) debiasing Glove embeddings. 
We observe a \SBB{significant} drop in correlation with gold scores, which is expected since debiasing is removing some information from the embeddings. 
In  spite of the drop, there is a reasonable correlation with the gold scores indicating that {\X} is able to retain a significant degree of semantic information. 


\begin{table}[t!]
    \centering
    \resizebox{0.48\textwidth}{!}{
		\begin{tabular}{ l c c c} 
			\toprule[1pt]
			{Method} & SimLex-999  & WordSim-353 & MTurk-771 \TBstrut\\
			\midrule[1pt]
            GloVe & 0.374 & {0.695} & {0.684} \\
            {\X} & 0.242 & 0.503 & 0.456\\
			\bottomrule[1pt]
	\end{tabular}
}
    \caption{\small Word similarity scores before and after debiasing GloVe embeddings using {\X}.}
    \vspace{-15pt}
    \label{tab:results-sim}
\end{table}

\noindent (b) \textit{Part-of-speech tagging}: We evaluate debiased embeddings 
for detecting POS tags in a sentence using the Universal tagset~\cite{petrov2011universal}. GloVe embeddings achieve an F1-score of 95.2\% and {\X} achieves an F1-score of 93.0\% on this task. This shows {\X}'s debiased embeddings still possess a  significant amount of morphological information about the language. 

\noindent (c) \textit{Sentiment Classification}: We perform sentiment classification using word embeddings on the IMDb movies dataset~\cite{maas2011learning}. 
GloVe embeddings achieve an accuracy of 80.9\%, while debiased embeddings achieve an accuracy of 74.6\%. The drop in this task is slightly more compared to POS tagging, but {\X} is still able to achieve reasonable performance on this task.

These experiments showcase that even though exact distances aren't preserved using {\X}, the debiased embeddings still retain relevant information useful in downstream tasks. 

\begin{figure}[t!]
    \centering
    
    \subfloat[][\footnotesize Unconstrained]{
    \resizebox{0.241\textwidth}{!}{
    \input{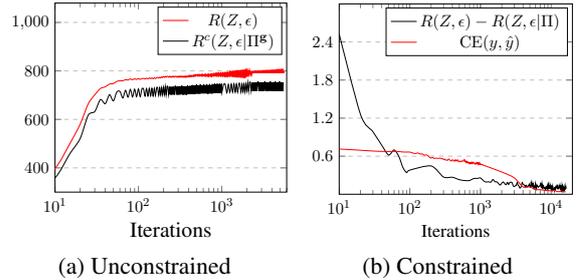}{\label{fig:loss-evolution-unconstrained}}
    }}
    \subfloat[][\footnotesize Constrained]{
    \resizebox{0.229\textwidth}{!}{
    \begin{tikzpicture}
    

\begin{axis}[
  ymin=0, ymax=3,
  xmin=10, xmax=20000,
  xtick={1, 10, 100, 1000, 10000},
  ytick={0.6, 1.2, 1.8, 2.4},
    ymajorgrids=true,
    grid style=dashed,
  xmode=log,
  xlabel={\large Iterations},
    scatter/classes={%
        a={mark=o,draw=black}}],
    
]
\addplot[smooth,mark=.,black]
  coordinates{
(0, 17.002200317382812)
(10, 2.5251108169555665)
(20, 1.2601482391357421)
(30, 0.9863201141357422)
(40, 0.7738286972045898)
(50, 0.6184309005737305)
(60, 0.6980560302734375)
(70, 0.5958627700805664)
(80, 0.4206699371337891)
(90, 0.34243431091308596)
(100, 0.3803455352783203)
(200, 0.4457719802856445)
(300, 0.2715452194213867)
(400, 0.26791973114013673)
(500, 0.23023300170898436)
(600, 0.21580352783203124)
(700, 0.21620140075683594)
(800, 0.2480073928833008)
(900, 0.24981002807617186)
(1000, 0.2571592330932617)
(1100, 0.21275672912597657)
(1200, 0.19234886169433593)
(1300, 0.19608573913574218)
(1400, 0.18974437713623046)
(1500, 0.23242568969726562)
(1600, 0.19770355224609376)
(1700, 0.19572696685791016)
(1800, 0.16826992034912108)
(1900, 0.1450735092163086)
(2000, 0.1750558853149414)
(2100, 0.19398860931396483)
(2200, 0.16656951904296874)
(2300, 0.1499462127685547)
(2400, 0.1864086151123047)
(2500, 0.15426292419433593)
(2600, 0.18814315795898437)
(2700, 0.18805084228515626)
(2800, 0.18332901000976562)
(2900, 0.13358230590820314)
(3000, 0.12090740203857422)
(3100, 0.1339792251586914)
(3200, 0.13773651123046876)
(3300, 0.18338069915771485)
(3400, 0.14316844940185547)
(3500, 0.13050460815429688)
(3600, 0.09761962890625)
(3700, 0.10587291717529297)
(3800, 0.14078121185302733)
(3900, 0.18358993530273438)
(4000, 0.1574399948120117)
(4100, 0.14974308013916016)
(4200, 0.14027271270751954)
(4300, 0.11742839813232422)
(4400, 0.14215431213378907)
(4500, 0.12311477661132812)
(4600, 0.1205606460571289)
(4700, 0.13316879272460938)
(4800, 0.1292041778564453)
(4900, 0.15974082946777343)
(5000, 0.12657470703125)
(5100, 0.11432571411132812)
(5200, 0.18892593383789064)
(5300, 0.12899379730224608)
(5400, 0.12520103454589843)
(5500, 0.1799398422241211)
(5600, 0.13362903594970704)
(5700, 0.09490089416503907)
(5800, 0.1254434585571289)
(5900, 0.10101394653320313)
(6000, 0.17670421600341796)
(6100, 0.12724075317382813)
(6200, 0.08591670989990234)
(6300, 0.14620475769042968)
(6400, 0.12715320587158202)
(6500, 0.11908226013183594)
(6600, 0.11074256896972656)
(6700, 0.12717247009277344)
(6800, 0.12984676361083985)
(6900, 0.1264944076538086)
(7000, 0.1410907745361328)
(7100, 0.16760692596435547)
(7200, 0.11871089935302734)
(7300, 0.1066436767578125)
(7400, 0.16748924255371095)
(7500, 0.09831161499023437)
(7600, 0.08317680358886718)
(7700, 0.11885280609130859)
(7800, 0.13752613067626954)
(7900, 0.10249652862548828)
(8000, 0.09495506286621094)
(8100, 0.10213184356689453)
(8200, 0.08195533752441406)
(8300, 0.07117729187011719)
(8400, 0.11761150360107422)
(8500, 0.11217842102050782)
(8600, 0.11212654113769531)
(8700, 0.0899282455444336)
(8800, 0.13319625854492187)
(8900, 0.0943267822265625)
(9000, 0.08484477996826172)
(9100, 0.14830417633056642)
(9200, 0.11163825988769531)
(9300, 0.1472332000732422)
(9400, 0.09674358367919922)
(9500, 0.1005819320678711)
(9600, 0.09497947692871093)
(9700, 0.09617767333984376)
(9800, 0.14177513122558594)
(9900, 0.11034126281738281)
(10000, 0.08579540252685547)
(10100, 0.06745662689208984)
(10200, 0.12424163818359375)
(10300, 0.10156307220458985)
(10400, 0.07110519409179687)
(10500, 0.13009490966796874)
(10600, 0.09804058074951172)
(10700, 0.06488094329833985)
(10800, 0.07283687591552734)
(10900, 0.08129348754882812)
(11000, 0.08258132934570313)
(11100, 0.05644073486328125)
(11200, 0.09687290191650391)
(11300, 0.10894317626953125)
(11400, 0.08488616943359376)
(11500, 0.07011337280273437)
(11600, 0.10959224700927735)
(11700, 0.12140216827392578)
(11800, 0.13405208587646483)
(11900, 0.11675453186035156)
(12000, 0.1091226577758789)
(12100, 0.09534854888916015)
(12200, 0.09332103729248047)
(12300, 0.13375701904296874)
(12400, 0.1456605911254883)
(12500, 0.12445220947265626)
(12600, 0.086944580078125)
(12700, 0.0969808578491211)
(12800, 0.0912851333618164)
(12900, 0.09594268798828125)
(13000, 0.11944770812988281)
(13100, 0.1323232650756836)
(13200, 0.1259153366088867)
(13300, 0.14524173736572266)
(13400, 0.10700454711914062)
(13500, 0.07947311401367188)
(13600, 0.09116382598876953)
(13700, 0.08683300018310547)
(13800, 0.1353616714477539)
(13900, 0.06901607513427735)
(14000, 0.07158088684082031)
(14100, 0.05507678985595703)
(14200, 0.05219764709472656)
(14300, 0.13854312896728516)
(14400, 0.11328582763671875)
(14500, 0.148291015625)
(14600, 0.047461318969726565)
(14700, 0.08917522430419922)
(14800, 0.061602783203125)
(14900, 0.05406169891357422)
(15000, 0.08961849212646485)
(15100, 0.06643695831298828)
(15200, 0.06727237701416015)
(15300, 0.10149402618408203)
(15400, 0.06387844085693359)
(15500, 0.06626014709472657)
(15600, 0.0980306625366211)
(15700, 0.08870792388916016)
(15800, 0.09027786254882812)
(15900, 0.1042083740234375)
(16000, 0.09489364624023437)
(16100, 0.12321453094482422)
(16200, 0.08316192626953126)
}; \addlegendentry{$R(Z, \epsilon) - R(Z, \epsilon|\Pi)$}

\addplot[smooth,mark=.,red]
  coordinates{
  (0, 0.7726842343807221)
(10, 0.7129248261451722)
(20, 0.6938621997833252)
(30, 0.6806760489940643)
(40, 0.6754227161407471)
(50, 0.6735133051872253)
(60, 0.6751242637634277)
(70, 0.6724383711814881)
(80, 0.6659818530082703)
(90, 0.6648479163646698)
(100, 0.6646653413772583)
(110, 0.6498369514942169)
(120, 0.6416367948055267)
(130, 0.6506009519100189)
(140, 0.633529680967331)
(150, 0.6368439614772796)
(160, 0.6163716614246368)
(170, 0.6192368865013123)
(180, 0.611096853017807)
(190, 0.5822762250900269)
(200, 0.5705192744731903)
(210, 0.58056560754776)
(220, 0.5650802433490754)
(230, 0.5683228433132171)
(240, 0.5683542549610138)
(250, 0.5641779869794845)
(260, 0.5673368394374847)
(270, 0.5686449587345124)
(280, 0.5598013997077942)
(290, 0.5616305112838745)
(300, 0.5480084419250488)
(310, 0.5544845640659333)
(320, 0.5388348877429963)
(330, 0.5411595523357391)
(340, 0.5328878730535507)
(350, 0.5497605711221695)
(360, 0.5541879713535309)
(370, 0.5367968916893006)
(380, 0.5281217634677887)
(390, 0.5253645241260528)
(400, 0.5442659318447113)
(410, 0.536802151799202)
(420, 0.5279710203409195)
(430, 0.5208738684654236)
(440, 0.543507969379425)
(450, 0.5282795250415802)
(460, 0.539487612247467)
(470, 0.5249476909637452)
(480, 0.5384453237056732)
(490, 0.5162836313247681)
(500, 0.5115217596292496)
(510, 0.5212237715721131)
(520, 0.5218010902404785)
(530, 0.5149543762207032)
(540, 0.5136588484048843)
(550, 0.5269901394844055)
(560, 0.5265650898218155)
(570, 0.5213458478450775)
(580, 0.5253670454025269)
(590, 0.5448617458343505)
(600, 0.5184650123119354)
(610, 0.5317160367965699)
(620, 0.5093383848667145)
(630, 0.496666219830513)
(640, 0.5167658656835556)
(650, 0.5026486545801163)
(660, 0.4972274720668793)
(670, 0.5091732829809189)
(680, 0.48696603775024416)
(690, 0.48716201186180114)
(700, 0.4978320121765137)
(710, 0.4820297956466675)
(720, 0.5000895380973815)
(730, 0.49681014716625216)
(740, 0.48606609404087064)
(750, 0.5066709965467453)
(760, 0.49279101490974425)
(770, 0.48345288038253786)
(780, 0.4932331621646881)
(790, 0.5033053159713745)
(800, 0.491570046544075)
(810, 0.4907878488302231)
(820, 0.4933394372463226)
(830, 0.47613735496997833)
(840, 0.47484728693962097)
(850, 0.47560938596725466)
(860, 0.4900978684425354)
(870, 0.49350749552249906)
(880, 0.47971844375133516)
(890, 0.4941446930170059)
(900, 0.5010710656642914)
(910, 0.4866110920906067)
(920, 0.49246775507926943)
(930, 0.4719048798084259)
(940, 0.48733638525009154)
(950, 0.49301197230815885)
(960, 0.48174501657485963)
(970, 0.4940839737653732)
(980, 0.4777127355337143)
(990, 0.47495557367801666)
(1000, 0.4744139552116394)
(2000, 0.35099715888500216)
(3000, 0.2591696515679359)
(4000, 0.11622141823172569)
(5000, 0.11491994485259056)
(6000, 0.0736666426062584)
(7000, 0.07271427065134048)
(8000, 0.06144315116107464)
(9000, 0.05736419670283795)
(10000, 0.05379610061645508)
(11000, 0.0611240316182375)
(12000, 0.0448768712580204)
(13000, 0.040413451939821245)
(14000, 0.04858638700097799)
(15000, 0.03603757619857788)
(16000, 0.040891601517796514)
}; \addlegendentry{$\mathrm{CE}(y, \hat{y})$}


\end{axis}
\end{tikzpicture}{\label{subfig:loss-evolution-constrained}}
    }}
    \vspace{-10pt}
    \caption{\small Loss evolution in the unconstained setup (left) where both terms -- $R(Z, \epsilon)$ (\textcolor{red}{\textbf{red}}) and $R^c(Z, \epsilon|\Pi^{\mathbf{g}})$ (\textcolor{black}{\textbf{black}}) start increasing simultaneously. In the constrained setup (right) with $\lambda = 0.01$ -- bias loss (\textbf{black}) starts converging earlier than the target loss (\textcolor{red}{\textbf{red}}). }
    \label{fig:loss-evolution}
    \vspace{-15pt}
\end{figure}

\noindent\textbf{Evolution of loss components.} We evaluate how {\X}'s loss components evolve during training. In the unconstrained setup for GloVe debiasing, we evaluate how the evolution of components -- $R(Z, \epsilon)$ (in \textcolor{red}{\textbf{red}}) and $R^c(Z, \epsilon|\Pi^{\mathbf{g}})$ (in \textcolor{black}{\textbf{black}}). In Figure~\ref{fig:loss-evolution-unconstrained}, we observe that both loss terms start increasing simultaneously, with their difference remaining constant in the final iterations. Next in the constrained setup, the evolution of target loss $\mathrm{CE}(\hat{y}, y)$  and bias loss  $R(Z, \epsilon) - R^c(Z, \epsilon|\Pi^{\mathbf{g}})$  for {\dial} dataset is shown in Figure~\ref{subfig:loss-evolution-constrained}. We observe that the bias term converges first followed by the target loss. This is expected as the magnitude of rate-distortion loss is larger than target loss, 
which forces the model to minimize it first. 

\noindent\textbf{Limitations.}\label{subsec:limitations} A limitation of {\X} is that we lack a principled feature map  $\phi(x)$ selection approach. In the unconstrained setup, we relied on empirical observations and found that a 4-layer ReLU network sufficed for GloVe and Biographies, while a 7-layer network was required for {\dial}. For the constrained setup, BERT\textsubscript{base} proved to be expressive enough to perform debiasing in all setups. Future works can explore white-box network architectures \cite{chan2021redunet} for debiasing.

\section{Conclusion}
We proposed {\FARM} ({\X}), a novel debiasing technique based on the principle of rate-distortion maximization. {\X} is effective in removing protected information from representations in both unconstrained and constrained debiasing setups. 
Empirical evaluations show that {\X} outperforms prior works in debiasing representations by a large margin on several datasets. Extensive analysis showcase that {\X} is sample efficient, and robust to label corruptions and minor hyperparameter changes. Future works can focus on leveraging {\X} for achieving fairness in complex tasks like language generation. 

\section*{\SBB{Ethical Considerations}}

\SBB{In this work, we present {\X} -- a robust representation learning framework to selectively remove protected information. {\X} is developed with an intent to enable development of fair learning systems. However, {\X} can be misused to remove salient features from representations and perform classification by leveraging demographic information. Debiasing using {\X} is only evaluated on datasets with binary protected attribute variables.
This may not be ideal while removing protected information about gender, which can extend beyond binary categories. Currently, we lack datasets with fine-grained gender annotation. It is important to collect data and develop techniques, that would benefit everyone in our community.}

\bibliography{tacl2021}

\begin{thebibliography}{61}
\expandafter\ifx\csname natexlab\endcsname\relax\def\natexlab#1{#1}\fi

\bibitem[{Agirre et~al.(2009{\natexlab{a}})Agirre, Alfonseca, Hall, Kravalova,
  Pa{\c{s}}ca, and Soroa}]{agirre2009study}
Eneko Agirre, Enrique Alfonseca, Keith Hall, Jana Kravalova, Marius
  Pa{\c{s}}ca, and Aitor Soroa. 2009{\natexlab{a}}.
\newblock \href {https://aclanthology.org/N09-1003} {A study on similarity and
  relatedness using distributional and {W}ord{N}et-based approaches}.
\newblock In \emph{Proceedings of Human Language Technologies: The 2009 Annual
  Conference of the North {A}merican Chapter of the Association for
  Computational Linguistics}, pages 19--27, Boulder, Colorado. Association for
  Computational Linguistics.

\bibitem[{Agirre et~al.(2009{\natexlab{b}})Agirre, Alfonseca, Hall, Kravalova,
  Pa{\c{s}}ca, and Soroa}]{agirre}
Eneko Agirre, Enrique Alfonseca, Keith Hall, Jana Kravalova, Marius
  Pa{\c{s}}ca, and Aitor Soroa. 2009{\natexlab{b}}.
\newblock \href {https://aclanthology.org/N09-1003} {A study on similarity and
  relatedness using distributional and {W}ord{N}et-based approaches}.
\newblock In \emph{Proceedings of Human Language Technologies: The 2009 Annual
  Conference of the North {A}merican Chapter of the Association for
  Computational Linguistics}, pages 19--27, Boulder, Colorado. Association for
  Computational Linguistics.

\bibitem[{Ba et~al.(2016)Ba, Kiros, and Hinton}]{ba2016layer}
Jimmy~Lei Ba, Jamie~Ryan Kiros, and Geoffrey~E Hinton. 2016.
\newblock Layer normalization.
\newblock \emph{arXiv preprint arXiv:1607.06450}.

\bibitem[{Barrett et~al.(2019)Barrett, Kementchedjhieva, Elazar, Elliott, and
  S{\o}gaard}]{barrett-etal-2019-adversarial}
Maria Barrett, Yova Kementchedjhieva, Yanai Elazar, Desmond Elliott, and Anders
  S{\o}gaard. 2019.
\newblock \href {https://doi.org/10.18653/v1/D19-1662} {Adversarial removal of
  demographic attributes revisited}.
\newblock In \emph{Proceedings of the 2019 Conference on Empirical Methods in
  Natural Language Processing and the 9th International Joint Conference on
  Natural Language Processing (EMNLP-IJCNLP)}, pages 6330--6335, Hong Kong,
  China. Association for Computational Linguistics.

\bibitem[{Basu et~al.(2019)Basu, Basu, Buckmire, and Lal}]{basu2019predictive}
Kanadpriya Basu, Treena Basu, Ron Buckmire, and Nishu Lal. 2019.
\newblock Predictive models of student college commitment decisions using
  machine learning.
\newblock \emph{Data}, 4(2):65.

\bibitem[{Basu Roy~Chowdhury et~al.(2021)Basu Roy~Chowdhury, Ghosh, Li, Oliva,
  Srivastava, and Chaturvedi}]{basu-roy-chowdhury-etal-2021-adversarial}
Somnath Basu Roy~Chowdhury, Sayan Ghosh, Yiyuan Li, Junier Oliva, Shashank
  Srivastava, and Snigdha Chaturvedi. 2021.
\newblock \href {https://doi.org/10.18653/v1/2021.emnlp-main.43} {Adversarial
  scrubbing of demographic information for text classification}.
\newblock In \emph{Proceedings of the 2021 Conference on Empirical Methods in
  Natural Language Processing}, pages 550--562, Online and Punta Cana,
  Dominican Republic. Association for Computational Linguistics.

\bibitem[{Bickel et~al.(1975)Bickel, Hammel, and O'Connell}]{bickel1975sex}
Peter~J Bickel, Eugene~A Hammel, and J~William O'Connell. 1975.
\newblock Sex bias in graduate admissions: Data from berkeley: Measuring bias
  is harder than is usually assumed, and the evidence is sometimes contrary to
  expectation.
\newblock \emph{Science}, 187(4175):398--404.

\bibitem[{Blodgett et~al.(2016)Blodgett, Green, and
  O{'}Connor}]{blodgett2016demographic}
Su~Lin Blodgett, Lisa Green, and Brendan O{'}Connor. 2016.
\newblock \href {https://doi.org/10.18653/v1/D16-1120} {Demographic dialectal
  variation in social media: A case study of {A}frican-{A}merican {E}nglish}.
\newblock In \emph{Proceedings of the 2016 Conference on Empirical Methods in
  Natural Language Processing}, pages 1119--1130, Austin, Texas. Association
  for Computational Linguistics.

\bibitem[{Bolukbasi et~al.(2016)Bolukbasi, Chang, Zou, Saligrama, and
  Kalai}]{bolukbasi2016man}
Tolga Bolukbasi, Kai{-}Wei Chang, James~Y. Zou, Venkatesh Saligrama, and
  Adam~Tauman Kalai. 2016.
\newblock \href
  {https://proceedings.neurips.cc/paper/2016/hash/a486cd07e4ac3d270571622f4f316ec5-Abstract.html}
  {Man is to computer programmer as woman is to homemaker? debiasing word
  embeddings}.
\newblock In \emph{Advances in Neural Information Processing Systems 29: Annual
  Conference on Neural Information Processing Systems 2016, December 5-10,
  2016, Barcelona, Spain}, pages 4349--4357.

\bibitem[{Bruni et~al.(2014)Bruni, Tran, and Baroni}]{bruni2014multimodal}
Elia Bruni, Nam-Khanh Tran, and Marco Baroni. 2014.
\newblock Multimodal distributional semantics.
\newblock \emph{Journal of artificial intelligence research}, 49:1--47.

\bibitem[{Burger et~al.(2011)Burger, Henderson, Kim, and
  Zarrella}]{burger2011discriminating}
John~D. Burger, John Henderson, George Kim, and Guido Zarrella. 2011.
\newblock \href {https://aclanthology.org/D11-1120} {Discriminating gender on
  {T}witter}.
\newblock In \emph{Proceedings of the 2011 Conference on Empirical Methods in
  Natural Language Processing}, pages 1301--1309, Edinburgh, Scotland, UK.
  Association for Computational Linguistics.

\bibitem[{Chan et~al.(2022)Chan, Yu, You, Qi, Wright, and Ma}]{chan2021redunet}
Kwan Ho~Ryan Chan, Yaodong Yu, Chong You, Haozhi Qi, John Wright, and Yi~Ma.
  2022.
\newblock \href {http://jmlr.org/papers/v23/21-0631.html} {Redunet: A white-box
  deep network from the principle of maximizing rate reduction}.
\newblock \emph{Journal of Machine Learning Research}, 23(114):1--103.

\bibitem[{Cover(1999)}]{cover1999elements}
Thomas~M Cover. 1999.
\newblock \emph{Elements of information theory}.
\newblock John Wiley \& Sons.

\bibitem[{De-Arteaga et~al.(2019)De-Arteaga, Romanov, Wallach, Chayes, Borgs,
  Chouldechova, Geyik, Kenthapadi, and Kalai}]{de2019bias}
Maria De-Arteaga, Alexey Romanov, Hanna Wallach, Jennifer Chayes, Christian
  Borgs, Alexandra Chouldechova, Sahin Geyik, Krishnaram Kenthapadi, and
  Adam~Tauman Kalai. 2019.
\newblock Bias in bios: A case study of semantic representation bias in a
  high-stakes setting.
\newblock In \emph{proceedings of the Conference on Fairness, Accountability,
  and Transparency}, pages 120--128.

\bibitem[{Dev et~al.(2021)Dev, Li, Phillips, and Srikumar}]{dev2020oscar}
Sunipa Dev, Tao Li, Jeff~M Phillips, and Vivek Srikumar. 2021.
\newblock \href {https://doi.org/10.18653/v1/2021.emnlp-main.411} {{OSC}a{R}:
  Orthogonal subspace correction and rectification of biases in word
  embeddings}.
\newblock In \emph{Proceedings of the 2021 Conference on Empirical Methods in
  Natural Language Processing}, pages 5034--5050, Online and Punta Cana,
  Dominican Republic. Association for Computational Linguistics.

\bibitem[{Devlin et~al.(2019)Devlin, Chang, Lee, and Toutanova}]{bert}
Jacob Devlin, Ming-Wei Chang, Kenton Lee, and Kristina Toutanova. 2019.
\newblock \href {https://doi.org/10.18653/v1/N19-1423} {{BERT}: Pre-training of
  deep bidirectional transformers for language understanding}.
\newblock In \emph{Proceedings of the 2019 Conference of the North {A}merican
  Chapter of the Association for Computational Linguistics: Human Language
  Technologies, Volume 1 (Long and Short Papers)}, pages 4171--4186,
  Minneapolis, Minnesota. Association for Computational Linguistics.

\bibitem[{Elazar and Goldberg(2018)}]{elazar2018adversarial}
Yanai Elazar and Yoav Goldberg. 2018.
\newblock \href {https://doi.org/10.18653/v1/D18-1002} {Adversarial removal of
  demographic attributes from text data}.
\newblock In \emph{Proceedings of the 2018 Conference on Empirical Methods in
  Natural Language Processing}, pages 11--21, Brussels, Belgium. Association
  for Computational Linguistics.

\bibitem[{Elazar et~al.(2021)Elazar, Ravfogel, Jacovi, and
  Goldberg}]{elazar2021amnesic}
Yanai Elazar, Shauli Ravfogel, Alon Jacovi, and Yoav Goldberg. 2021.
\newblock Amnesic probing: Behavioral explanation with amnesic counterfactuals.
\newblock \emph{Transactions of the Association for Computational Linguistics},
  9:160--175.

\bibitem[{Felbo et~al.(2017)Felbo, Mislove, S{\o}gaard, Rahwan, and
  Lehmann}]{felbo2017using}
Bjarke Felbo, Alan Mislove, Anders S{\o}gaard, Iyad Rahwan, and Sune Lehmann.
  2017.
\newblock \href {https://doi.org/10.18653/v1/D17-1169} {Using millions of emoji
  occurrences to learn any-domain representations for detecting sentiment,
  emotion and sarcasm}.
\newblock In \emph{Proceedings of the 2017 Conference on Empirical Methods in
  Natural Language Processing}, pages 1615--1625, Copenhagen, Denmark.
  Association for Computational Linguistics.

\bibitem[{Finkelstein et~al.(2001)Finkelstein, Gabrilovich, Matias, Rivlin,
  Solan, Wolfman, and Ruppin}]{finkelstein2001placing}
Lev Finkelstein, Evgeniy Gabrilovich, Yossi Matias, Ehud Rivlin, Zach Solan,
  Gadi Wolfman, and Eytan Ruppin. 2001.
\newblock Placing search in context: The concept revisited.
\newblock In \emph{Proceedings of the 10th international conference on World
  Wide Web}, pages 406--414.

\bibitem[{Fleming and Jamison(2003)}]{jamison2002isometries}
Richard~J Fleming and James~E Jamison. 2003.
\newblock \emph{Function Spaces}.
\newblock Chapman \& Hall/CRC.

\bibitem[{Gerz et~al.(2016)Gerz, Vuli{\'c}, Hill, Reichart, and
  Korhonen}]{gerz}
Daniela Gerz, Ivan Vuli{\'c}, Felix Hill, Roi Reichart, and Anna Korhonen.
  2016.
\newblock \href {https://doi.org/10.18653/v1/D16-1235} {{S}im{V}erb-3500: A
  large-scale evaluation set of verb similarity}.
\newblock In \emph{Proceedings of the 2016 Conference on Empirical Methods in
  Natural Language Processing}, pages 2173--2182, Austin, Texas. Association
  for Computational Linguistics.

\bibitem[{Ghailan et~al.(2016)Ghailan, Mokhtar, and
  Hegazy}]{ghailan2016improving}
Omar Ghailan, Hoda~MO Mokhtar, and Osman Hegazy. 2016.
\newblock Improving credit scorecard modeling through applying text analysis.
\newblock \emph{institutions}, 7(4).

\bibitem[{Goodfellow et~al.(2014)Goodfellow, Pouget-Abadie, Mirza, Xu,
  Warde-Farley, Ozair, Courville, and Bengio}]{goodfellow2014generative}
Ian Goodfellow, Jean Pouget-Abadie, Mehdi Mirza, Bing Xu, David Warde-Farley,
  Sherjil Ozair, Aaron Courville, and Yoshua Bengio. 2014.
\newblock \href
  {https://proceedings.neurips.cc/paper/2014/file/5ca3e9b122f61f8f06494c97b1afccf3-Paper.pdf}
  {Generative adversarial nets}.
\newblock In \emph{Advances in Neural Information Processing Systems},
  volume~27. Curran Associates, Inc.

\bibitem[{Halawi et~al.(2012)Halawi, Dror, Gabrilovich, and
  Koren}]{halawi2012large}
Guy Halawi, Gideon Dror, Evgeniy Gabrilovich, and Yehuda Koren. 2012.
\newblock \href {https://doi.org/10.1145/2339530.2339751} {Large-scale learning
  of word relatedness with constraints}.
\newblock In \emph{The 18th {ACM} {SIGKDD} International Conference on
  Knowledge Discovery and Data Mining, {KDD} '12, Beijing, China, August 12-16,
  2012}, pages 1406--1414. {ACM}.

\bibitem[{Hill et~al.(2015)Hill, Reichart, and Korhonen}]{hill2015simlex}
Felix Hill, Roi Reichart, and Anna Korhonen. 2015.
\newblock \href {https://doi.org/10.1162/COLI_a_00237} {{S}im{L}ex-999:
  Evaluating semantic models with (genuine) similarity estimation}.
\newblock \emph{Computational Linguistics}, 41(4):665--695.

\bibitem[{Huang et~al.(2012)Huang, Socher, Manning, and
  Ng}]{huang2012improving}
Eric~H Huang, Richard Socher, Christopher~D Manning, and Andrew~Y Ng. 2012.
\newblock Improving word representations via global context and multiple word
  prototypes.
\newblock In \emph{Proceedings of the 50th Annual Meeting of the Association
  for Computational Linguistics (Volume 1: Long Papers)}, pages 873--882.

\bibitem[{Joulin et~al.(2017)Joulin, Grave, Bojanowski, and Mikolov}]{fasttext}
Armand Joulin, Edouard Grave, Piotr Bojanowski, and Tomas Mikolov. 2017.
\newblock \href {https://aclanthology.org/E17-2068} {Bag of tricks for
  efficient text classification}.
\newblock In \emph{Proceedings of the 15th Conference of the {E}uropean Chapter
  of the Association for Computational Linguistics: Volume 2, Short Papers},
  pages 427--431, Valencia, Spain. Association for Computational Linguistics.

\bibitem[{Koppel et~al.(2002)Koppel, Argamon, and
  Shimoni}]{koppel2002automatically}
Moshe Koppel, Shlomo Argamon, and Anat~Rachel Shimoni. 2002.
\newblock Automatically categorizing written texts by author gender.
\newblock \emph{Literary and linguistic computing}, 17(4):401--412.

\bibitem[{Li et~al.(2018)Li, Baldwin, and Cohn}]{li2018towards}
Yitong Li, Timothy Baldwin, and Trevor Cohn. 2018.
\newblock \href {https://doi.org/10.18653/v1/P18-2005} {Towards robust and
  privacy-preserving text representations}.
\newblock In \emph{Proceedings of the 56th Annual Meeting of the Association
  for Computational Linguistics (Volume 2: Short Papers)}, pages 25--30,
  Melbourne, Australia. Association for Computational Linguistics.

\bibitem[{Loshchilov and Hutter(2019)}]{loshchilov2017decoupled}
Ilya Loshchilov and Frank Hutter. 2019.
\newblock \href {https://openreview.net/forum?id=Bkg6RiCqY7} {Decoupled weight
  decay regularization}.
\newblock In \emph{7th International Conference on Learning Representations,
  {ICLR} 2019, New Orleans, LA, USA, May 6-9, 2019}. OpenReview.net.

\bibitem[{Luong et~al.(2013)Luong, Socher, and Manning}]{luong2013better}
Minh-Thang Luong, Richard Socher, and Christopher~D Manning. 2013.
\newblock Better word representations with recursive neural networks for
  morphology.
\newblock In \emph{Proceedings of the seventeenth conference on computational
  natural language learning}, pages 104--113.

\bibitem[{Ma et~al.(2007)Ma, Derksen, Hong, and Wright}]{ma2007segmentation}
Yi~Ma, Harm Derksen, Wei Hong, and John Wright. 2007.
\newblock \href {https://doi.org/10.1109/TPAMI.2007.1085} {Segmentation of
  multivariate mixed data via lossy data coding and compression}.
\newblock \emph{IEEE Transactions on Pattern Analysis and Machine
  Intelligence}, 29(9):1546--1562.

\bibitem[{Maas et~al.(2011)Maas, Daly, Pham, Huang, Ng, and
  Potts}]{maas2011learning}
Andrew~L. Maas, Raymond~E. Daly, Peter~T. Pham, Dan Huang, Andrew~Y. Ng, and
  Christopher Potts. 2011.
\newblock \href {https://aclanthology.org/P11-1015} {Learning word vectors for
  sentiment analysis}.
\newblock In \emph{Proceedings of the 49th Annual Meeting of the Association
  for Computational Linguistics: Human Language Technologies}, pages 142--150,
  Portland, Oregon, USA. Association for Computational Linguistics.

\bibitem[{Van~der Maaten and Hinton(2008)}]{van2008visualizing}
Laurens Van~der Maaten and Geoffrey Hinton. 2008.
\newblock Visualizing data using t-sne.
\newblock \emph{Journal of machine learning research}, 9(11).

\bibitem[{Macdonald et~al.(2019)Macdonald, W{\"a}ldchen, Hauch, and
  Kutyniok}]{macdonald2019rate}
Jan Macdonald, Stephan W{\"a}ldchen, Sascha Hauch, and Gitta Kutyniok. 2019.
\newblock \href {https://arxiv.org/abs/1905.11092} {A rate-distortion framework
  for explaining neural network decisions}.
\newblock \emph{ArXiv preprint}, abs/1905.11092.

\bibitem[{Mehrabi et~al.(2021)Mehrabi, Morstatter, Saxena, Lerman, and
  Galstyan}]{mehrabi2019survey}
Ninareh Mehrabi, Fred Morstatter, Nripsuta Saxena, Kristina Lerman, and Aram
  Galstyan. 2021.
\newblock \href {https://doi.org/10.1145/3457607} {A survey on bias and
  fairness in machine learning}.
\newblock \emph{ACM Comput. Surv.}, 54(6).

\bibitem[{Miller and Charles(1991)}]{mc_28}
George~A. Miller and Walter~G. Charles. 1991.
\newblock \href {https://doi.org/10.1080/01690969108406936} {Contextual
  correlates of semantic similarity}.
\newblock \emph{Language and Cognitive Processes}, 6(1):1--28.

\bibitem[{Nguyen et~al.(2013)Nguyen, Gravel, Trieschnigg, and
  Meder}]{nguyen2013old}
Dong Nguyen, Rilana Gravel, Dolf Trieschnigg, and Theo Meder. 2013.
\newblock " how old do you think i am?" a study of language and age in twitter.
\newblock In \emph{Proceedings of the International AAAI Conference on Web and
  Social Media}, volume~7.

\bibitem[{Pedregosa et~al.(2011)Pedregosa, Varoquaux, Gramfort, Michel,
  Thirion, Grisel, Blondel, Prettenhofer, Weiss, Dubourg, Vanderplas, Passos,
  Cournapeau, Brucher, Perrot, and {{\'E}}douard
  Duchesnay}]{pedregosa2011scikit}
Fabian Pedregosa, Ga{{\"e}}l Varoquaux, Alexandre Gramfort, Vincent Michel,
  Bertrand Thirion, Olivier Grisel, Mathieu Blondel, Peter Prettenhofer, Ron
  Weiss, Vincent Dubourg, Jake Vanderplas, Alexandre Passos, David Cournapeau,
  Matthieu Brucher, Matthieu Perrot, and {{\'E}}douard Duchesnay. 2011.
\newblock \href {http://jmlr.org/papers/v12/pedregosa11a.html} {Scikit-learn:
  Machine learning in python}.
\newblock \emph{Journal of Machine Learning Research}, 12(85):2825--2830.

\bibitem[{Petrov et~al.(2012)Petrov, Das, and McDonald}]{petrov2011universal}
Slav Petrov, Dipanjan Das, and Ryan McDonald. 2012.
\newblock \href
  {http://www.lrec-conf.org/proceedings/lrec2012/pdf/274_Paper.pdf} {A
  universal part-of-speech tagset}.
\newblock In \emph{Proceedings of the Eighth International Conference on
  Language Resources and Evaluation ({LREC}'12)}, pages 2089--2096, Istanbul,
  Turkey. European Language Resources Association (ELRA).

\bibitem[{Pirr{\'o} and Seco(2008)}]{pirro2008design}
Giuseppe Pirr{\'o} and Nuno Seco. 2008.
\newblock Design, implementation and evaluation of a new semantic similarity
  metric combining features and intrinsic information content.
\newblock In \emph{OTM Confederated International Conferences" On the Move to
  Meaningful Internet Systems"}, pages 1271--1288. Springer.

\bibitem[{Radinsky et~al.(2011)Radinsky, Agichtein, Gabrilovich, and
  Markovitch}]{radinsky2011word}
Kira Radinsky, Eugene Agichtein, Evgeniy Gabrilovich, and Shaul Markovitch.
  2011.
\newblock A word at a time: computing word relatedness using temporal semantic
  analysis.
\newblock In \emph{Proceedings of the 20th international conference on World
  wide web}, pages 337--346.

\bibitem[{Rangel et~al.(2016)Rangel, Rosso, Verhoeven, Daelemans, Potthast, and
  Stein}]{rangel2016overview}
Francisco Rangel, Paolo Rosso, Ben Verhoeven, Walter Daelemans, Martin
  Potthast, and Benno Stein. 2016.
\newblock Overview of the 4th author profiling task at pan 2016: cross-genre
  evaluations.
\newblock \emph{Working Notes Papers of the CLEF}, 2016:750--784.

\bibitem[{Ravfogel et~al.(2020)Ravfogel, Elazar, Gonen, Twiton, and
  Goldberg}]{ravfogel2020null}
Shauli Ravfogel, Yanai Elazar, Hila Gonen, Michael Twiton, and Yoav Goldberg.
  2020.
\newblock \href {https://doi.org/10.18653/v1/2020.acl-main.647} {Null it out:
  Guarding protected attributes by iterative nullspace projection}.
\newblock In \emph{Proceedings of the 58th Annual Meeting of the Association
  for Computational Linguistics}, pages 7237--7256, Online. Association for
  Computational Linguistics.

\bibitem[{Romanov et~al.(2019)Romanov, De-Arteaga, Wallach, Chayes, Borgs,
  Chouldechova, Geyik, Kenthapadi, Rumshisky, and Kalai}]{biasbios2}
Alexey Romanov, Maria De-Arteaga, Hanna Wallach, Jennifer Chayes, Christian
  Borgs, Alexandra Chouldechova, Sahin Geyik, Krishnaram Kenthapadi, Anna
  Rumshisky, and Adam Kalai. 2019.
\newblock \href {https://doi.org/10.18653/v1/N19-1424} {What{'}s in a name?
  {R}educing bias in bios without access to protected attributes}.
\newblock In \emph{Proceedings of the 2019 Conference of the North {A}merican
  Chapter of the Association for Computational Linguistics: Human Language
  Technologies, Volume 1 (Long and Short Papers)}, pages 4187--4195,
  Minneapolis, Minnesota. Association for Computational Linguistics.

\bibitem[{Rosenberg and Hirschberg(2007)}]{v-measure}
Andrew Rosenberg and Julia Hirschberg. 2007.
\newblock \href {https://aclanthology.org/D07-1043} {{V}-measure: A conditional
  entropy-based external cluster evaluation measure}.
\newblock In \emph{Proceedings of the 2007 Joint Conference on Empirical
  Methods in Natural Language Processing and Computational Natural Language
  Learning ({EMNLP}-{C}o{NLL})}, pages 410--420, Prague, Czech Republic.
  Association for Computational Linguistics.

\bibitem[{Rubenstein and Goodenough(1965)}]{rubenstein1965contextual}
Herbert Rubenstein and John~B Goodenough. 1965.
\newblock Contextual correlates of synonymy.
\newblock \emph{Communications of the ACM}, 8(10):627--633.

\bibitem[{Subramanian et~al.(2021)Subramanian, Han, Baldwin, Cohn, and
  Frermann}]{subramanian2021evaluating}
Shivashankar Subramanian, Xudong Han, Timothy Baldwin, Trevor Cohn, and Lea
  Frermann. 2021.
\newblock \href {https://doi.org/10.18653/v1/2021.emnlp-main.193} {Evaluating
  debiasing techniques for intersectional biases}.
\newblock In \emph{Proceedings of the 2021 Conference on Empirical Methods in
  Natural Language Processing}, pages 2492--2498, Online and Punta Cana,
  Dominican Republic. Association for Computational Linguistics.

\bibitem[{Szumlanski et~al.(2013)Szumlanski, Gomez, and
  Sims}]{szumlanski2013new}
Sean Szumlanski, Fernando Gomez, and Valerie~K Sims. 2013.
\newblock A new set of norms for semantic relatedness measures.
\newblock In \emph{Proceedings of the 51st Annual Meeting of the Association
  for Computational Linguistics (Volume 2: Short Papers)}, pages 890--895.

\bibitem[{Verhoeven and Daelemans(2014)}]{verhoeven2014clips}
Ben Verhoeven and Walter Daelemans. 2014.
\newblock \href {http://www.lrec-conf.org/proceedings/lrec2014/pdf/1_Paper.pdf}
  {{CL}i{PS} stylometry investigation ({CSI}) corpus: A {D}utch corpus for the
  detection of age, gender, personality, sentiment and deception in text}.
\newblock In \emph{Proceedings of the Ninth International Conference on
  Language Resources and Evaluation ({LREC}'14)}, pages 3081--3085, Reykjavik,
  Iceland. European Language Resources Association (ELRA).

\bibitem[{Verhoeven et~al.(2016)Verhoeven, Daelemans, and
  Plank}]{verhoeven2016twisty}
Ben Verhoeven, Walter Daelemans, and Barbara Plank. 2016.
\newblock \href {https://aclanthology.org/L16-1258} {{T}wi{S}ty: A multilingual
  {T}witter stylometry corpus for gender and personality profiling}.
\newblock In \emph{Proceedings of the Tenth International Conference on
  Language Resources and Evaluation ({LREC}'16)}, pages 1632--1637,
  Portoro{\v{z}}, Slovenia. European Language Resources Association (ELRA).

\bibitem[{Voita and Titov(2020)}]{voita2020information}
Elena Voita and Ivan Titov. 2020.
\newblock \href {https://doi.org/10.18653/v1/2020.emnlp-main.14}
  {Information-theoretic probing with minimum description length}.
\newblock In \emph{Proceedings of the 2020 Conference on Empirical Methods in
  Natural Language Processing (EMNLP)}, pages 183--196, Online. Association for
  Computational Linguistics.

\bibitem[{Weren et~al.(2014)Weren, Kauer, Mizusaki, Moreira, de~Oliveira, and
  Wives}]{weren2014examining}
Edson~RD Weren, Anderson~U Kauer, Lucas Mizusaki, Viviane~P Moreira,
  J~Palazzo~M de~Oliveira, and Leandro~K Wives. 2014.
\newblock Examining multiple features for author profiling.
\newblock \emph{Journal of information and data management}, 5(3):266--266.

\bibitem[{Xie et~al.(2017)Xie, Dai, Du, Hovy, and Neubig}]{xie2017controllable}
Qizhe Xie, Zihang Dai, Yulun Du, Eduard~H. Hovy, and Graham Neubig. 2017.
\newblock \href
  {https://proceedings.neurips.cc/paper/2017/hash/8cb22bdd0b7ba1ab13d742e22eed8da2-Abstract.html}
  {Controllable invariance through adversarial feature learning}.
\newblock In \emph{Advances in Neural Information Processing Systems 30: Annual
  Conference on Neural Information Processing Systems 2017, December 4-9, 2017,
  Long Beach, CA, {USA}}, pages 585--596.

\bibitem[{Yang and Powers(2006)}]{yang2006verb}
Dongqiang Yang and David~MW Powers. 2006.
\newblock \emph{Verb similarity on the taxonomy of WordNet}.
\newblock Masaryk University.

\bibitem[{Yu et~al.(2020)Yu, Chan, You, Song, and Ma}]{yu2020learning}
Yaodong Yu, Kwan Ho~Ryan Chan, Chong You, Chaobing Song, and Yi~Ma. 2020.
\newblock \href
  {https://proceedings.neurips.cc/paper/2020/file/6ad4174eba19ecb5fed17411a34ff5e6-Paper.pdf}
  {Learning diverse and discriminative representations via the principle of
  maximal coding rate reduction}.
\newblock In \emph{Advances in Neural Information Processing Systems},
  volume~33, pages 9422--9434. Curran Associates, Inc.

\bibitem[{Zemel et~al.(2013)Zemel, Wu, Swersky, Pitassi, and
  Dwork}]{zemel2013learning}
Richard~S. Zemel, Yu~Wu, Kevin Swersky, Toniann Pitassi, and Cynthia Dwork.
  2013.
\newblock \href {http://proceedings.mlr.press/v28/zemel13.html} {Learning fair
  representations}.
\newblock In \emph{Proceedings of the 30th International Conference on Machine
  Learning, {ICML} 2013, Atlanta, GA, USA, 16-21 June 2013}, volume~28 of
  \emph{{JMLR} Workshop and Conference Proceedings}, pages 325--333. JMLR.org.

\bibitem[{Zhang et~al.(2018)Zhang, Lemoine, and Mitchell}]{zhang2018mitigating}
Brian~Hu Zhang, Blake Lemoine, and Margaret Mitchell. 2018.
\newblock Mitigating unwanted biases with adversarial learning.
\newblock In \emph{Proceedings of the 2018 AAAI/ACM Conference on AI, Ethics,
  and Society}, pages 335--340.

\bibitem[{Zhao and Gordon(2019)}]{zhao2019inherent}
Han Zhao and Geoffrey~J. Gordon. 2019.
\newblock \href
  {https://proceedings.neurips.cc/paper/2019/hash/b4189d9de0fb2b9cce090bd1a15e3420-Abstract.html}
  {Inherent tradeoffs in learning fair representations}.
\newblock In \emph{Advances in Neural Information Processing Systems 32: Annual
  Conference on Neural Information Processing Systems 2019, NeurIPS 2019,
  December 8-14, 2019, Vancouver, BC, Canada}, pages 15649--15659.

\bibitem[{Zhao et~al.(2018)Zhao, Zhou, Li, Wang, and Chang}]{zhao2018learning}
Jieyu Zhao, Yichao Zhou, Zeyu Li, Wei Wang, and Kai-Wei Chang. 2018.
\newblock \href {https://doi.org/10.18653/v1/D18-1521} {Learning gender-neutral
  word embeddings}.
\newblock In \emph{Proceedings of the 2018 Conference on Empirical Methods in
  Natural Language Processing}, pages 4847--4853, Brussels, Belgium.
  Association for Computational Linguistics.

\end{thebibliography}
\bibliographystyle{acl_natbib}

    \appendix
\section{Appendix}






\subsection{Ablation: Unconstrained Debiasing}
\label{sec:ablation}
In this section, we present an ablation of {\X} where we utilize protected information loss of the unconstrained setup to mimic constrained debiasing. The modified objective function for unconstrained debiasing is shown as:
\begin{equation}
    \mathcal{L}_U(Z, \Pi^g) = \big[R^c(Z, \epsilon|\Pi^\mathbf{g}) - R(Z, \epsilon)\big]
\end{equation}

We evaluate this objective function for word embedding debiasing task. We obtain gender prediction accuracy of 49.8\% from the debiased embeddings, which slightly is better than the reported results in Table~\ref{tab:results-glove}. However, rank of the debiased embedding matrix is 2 (original dimension: 300), which shows that most of the information (including gender) has been destroyed during debiasing. This shows the importance of maximizing the overall rate-distortion term $R(Z, \epsilon)$, which helps in retaining diverse information from the original embedding space.


\subsection{Word Similarity Scores}
\label{sec:word-sim}
We evaluate the debiased word embeddings retrieved from {\X} on a large set of word similarity datasets. In Table~\ref{tab:similarity}, we report the similarity scores compared to GloVe embeddings. In all datasets, we observe a significant drop in correlation with the similarity scores, which is expected as debiasing using {\X} does not retain the subspace structure of the representations. We observe that debiased word embeddings from {\X} still retains a significant degree of semantic information, and can be useful in downstream tasks as demonstrated in Section~\ref{sec:probing-we}.
    
\begin{table}[t!]
    \centering
    \resizebox{0.5\textwidth}{!}{
\begin{tabular}{l|c c}
    \toprule[1pt]
    Dataset & GloVe & FaRM  \\
    \midrule[0.5pt]
     SimLex-999 \citep{hill2015simlex} & 0.374 & 0.242\\ 
     WordSim-353 \citep{finkelstein2001placing} & 0.695 & 0.503 \\
     MTurk-711 \citep{halawi2012large} & 0.684 & 0.456 \\
     Aggire \citep{agirre} & 0.690 & 0.480\\
     SimVerb-3500 \citep{gerz} & 0.226 & 0.140\\
     MEN \citep{bruni2014multimodal} & 0.731 & 0.525\\
     MC-28 \citep{mc_28} & 0.856 & 0.585\\
     PirroSeco \citep{pirro2008design} & 0.789 & 0.472\\
     MTurk-287 \citep{radinsky2011word} & 0.688 & 0.488\\
     RareWords 1401 & 0.414 & 0.272 \\
     RareWords 2304 \citep{luong2013better} & 0.385 & 0.245\\
     Rel 122 \citep{szumlanski2013new} & 0.687 & 0.395\\
     Rubenstien (\citeauthor{rubenstein1965contextual}) & 0.817 & 0.488\\
     SCWS \citep{huang2012improving} & 0.540 & 0.451\\
     SimLex-111 \citep{hill2015simlex} & 0.574 & 0.505\\
     YP-130 \citep{yang2006verb} & 0.537 & 0.442\\
     \bottomrule[1pt]
\end{tabular}
}
    \caption{Word similarity scores before and after debiasing GloVe embeddings using FaRM.}
    \label{tab:similarity}
\end{table}

\end{document}